\documentclass[acmtog]{acmart} 

\usepackage{epsfig}
\usepackage{graphicx}
\usepackage{amsmath}
\usepackage{graphicx}
\usepackage{mathtools}
\usepackage{subcaption}
\usepackage{tikz}
\usepackage{xspace}
\usepackage{xcolor}

\definecolor{our-orange}{HTML}{FFC348}
\definecolor{our-blue}{HTML}{BCD8FF}
\definecolor{our-dark-grey}{HTML}{4D4D4D}
\definecolor{our-light-grey}{HTML}{C6C6C6}

\definecolor{mib-comment-color}{HTML}{CC3333}

\AtBeginDocument{%
  \providecommand\BibTeX{{%
    \normalfont B\kern-0.5em{\scshape i\kern-0.25em b}\kern-0.8em\TeX}}}

\setcopyright{acmcopyright}
\copyrightyear{2022}
\acmYear{2022}
\acmDOI{XXXXXXX.XXXXXXX}

\acmConference[Conference acronym 'XX]{Make sure to enter the correct
  conference title from your rights confirmation emai}{June 03--05,
  2018}{Woodstock, NY}
\acmPrice{15.00}
\acmISBN{978-1-4503-XXXX-X/18/06}

\newcommand{\osm}{\textsc{OSM}\xspace}
\newcommand{\cmod}{\textsc{CompressionModel}\xspace}
\newcommand{\imod}{\textsc{IndexModel}\xspace}
\newcommand{\cpara}[1]{\noindent \textbf{#1} }

\begin{document}

\title{Large-Scale Auto-Regressive Modeling Of Street Networks}

\author{Michael Birsak}
\affiliation{%
  \institution{KAUST}
  \country{Kingdom of Saudi Arabia}
}

\author{Tom Kelly}
\affiliation{%
  \institution{KAUST}
  \country{Kingdom of Saudi Arabia}
}

\author{Wamiq Para}
\affiliation{%
  \institution{KAUST}
  \country{Kingdom of Saudi Arabia}
}

\author{Peter Wonka}
\affiliation{%
  \institution{KAUST}
  \country{Kingdom of Saudi Arabia}
}

\renewcommand{\shortauthors}{Birsak, et al.}

\begin{abstract}
    We present a novel generative method for the creation of city-scale road layouts. While the output of recent methods is limited in both size of the covered area and diversity, our framework produces large traversable graphs of high quality consisting of vertices and edges representing complete street networks covering 400 km² or more. While our framework can process general 2D embedded graphs, we focus on street networks due to the wide availability of training data.
    
    Our generative framework consists of a transformer decoder that is used in a sliding window manner to predict a field of indices, with each index encoding a representation of the local neighborhood. The semantics of each index is determined by a dictionary of context vectors. The index field is then input to a decoder to compute the street graph.
    Using data from OpenStreetMap, we train our system on whole cities and even across large countries such as the US, and finally compare it to the state of the art.
\end{abstract}
\begin{teaserfigure}
    \centering
    \includegraphics[width=1.0\columnwidth]{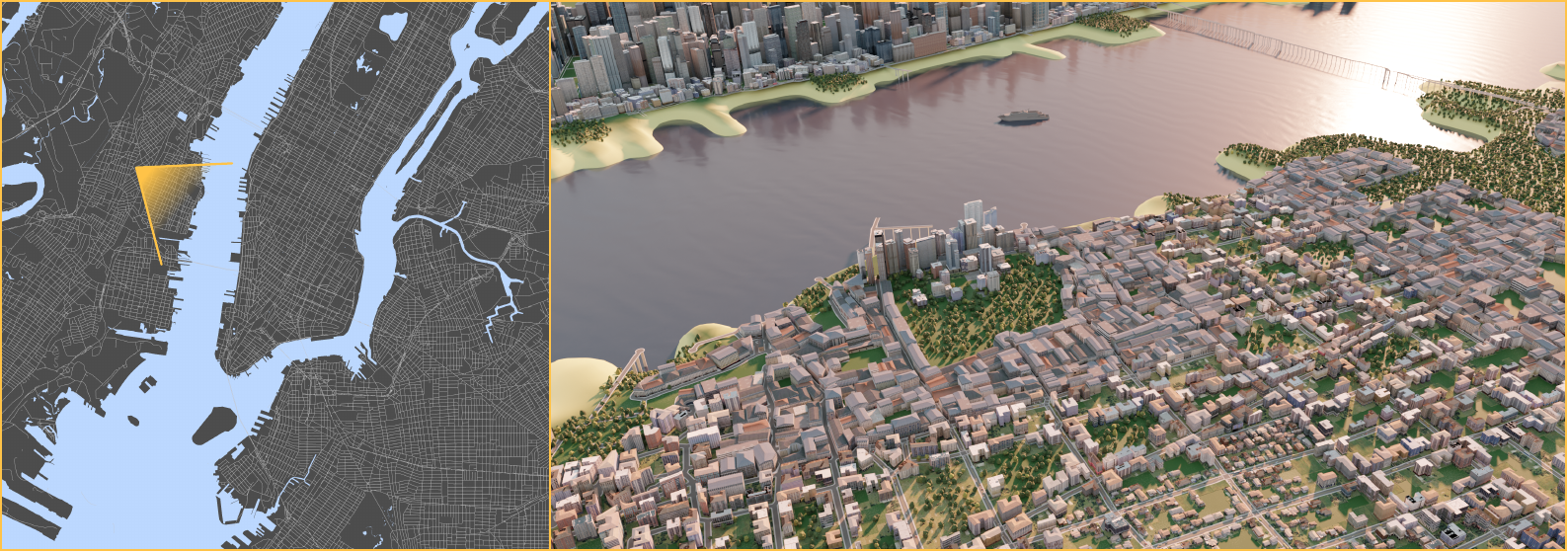}
    \caption{Our system creates diverse street networks at city-scale. Left: A generated layout over an area of 400 km² (20km x 20km) and view frustum (yellow triangle). Right: A rendering of the street graph from the view frustum.}
    \label{fig:teaser}
\end{teaserfigure}

 %



\maketitle

\section{Introduction}
Generative modeling is an active field of research with many exciting recent contributions. Examples include the generation of realistic human faces using the StyleGAN architecture~\cite{Karras2020ada,karras2021aliasfree} and the synthesis of images from language prompts~\cite{ramesh2021zeroshot,DallE2}.
The goal of this work is to explore the generative modeling of embedded graphs. In particular, we focus on embedded street graphs where each vertex is associated with a 2D coordinate.

We improve on a body of existing work in generative modeling; a typical example of which is Neural Turtle Graphics~\cite{chu2019ntg}, which gives high-quality results for small regions using GRU cells~\cite{li2016gated}. However, our analysis shows three shortcomings of this work that we address. First, the approach has an inherent limitation in that it encodes and processes graphs as a collection of polylines rather than considering all relevant information (e.g., patterns in the local neighborhood). Even locally, the model does not have a full understanding of its neighborhood. Second, the approach is limited to small areas; for city-scale street networks it is important to scale up the area generated by at least two orders of magnitude. Third, GRU cells have been surpassed by contemporary state-of-the-art generative models, suggesting that more recent approaches may produce stronger results. To overcome these limitations, we propose the following ideas.

\begin{figure*}
    \centering
         \def\svgwidth{1\linewidth}
        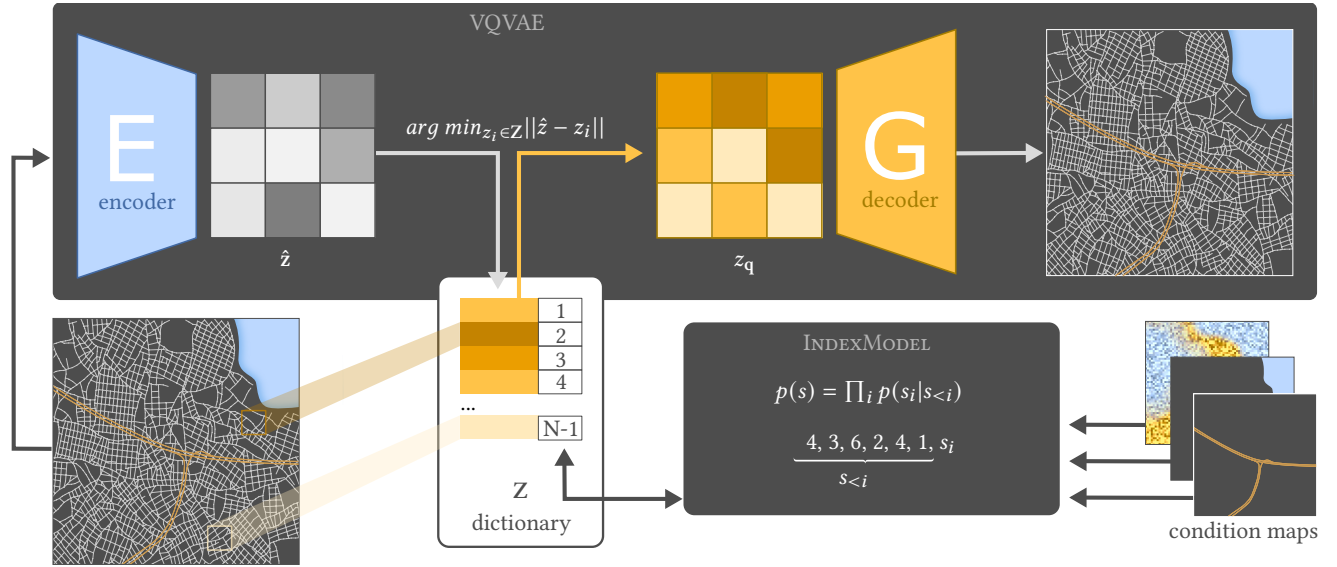 
    \caption{We leverage a VQVAE to obtain a context-rich dictionary. A transformer decoder referred to as the \imod is trained to learn a conditional prior of the index distribution. The decoder part of the VQVAE together with a thinning and graph tracing approach proposed in \cite{roadtracer} is used to obtain a traversable graph.}
    \label{fig:overview}
\end{figure*}


First, instead of GRU cells we use a transformer architecture~\cite{VaswaniAttention} which is a state-of-the-art architecture for generative modeling.
Second, we build on auto-regressive models that operate in a compressed space computed by trainable vector quantization~\cite{OordVAE}. Working in a compressed space provides better scalability of the model since a large-scale transformer architecture is very hard to train and requires a significant amount of hardware.
Third, we use a CNN-based decoder to decompress such a quantized representation into an image-based representation. Finally, we use a thinning approach together with the road-tracing approach proposed in \cite{roadtracer} to obtain a street graph. 
Fourth, we employ a combination of sliding window generation and decoding together with conditioning to generate coherent large-scale results. A standard sliding window approach alone would lead to incoherent results.
An overview of our architecture is shown in Figure~\ref{fig:overview}.

In summary, our paper contributes a large-scale modeling framework for high-quality street networks that is able to generate street networks that are two orders of magnitude larger than previous generative models for 2D embedded graphs including street graphs.
\vspace{-0.3cm}

\section{Related Work}

\subsection{Classical Approaches}

Street networks form a key element of generating urban environments. Traditional modeling approaches have been largely \emph{procedural} or rule-based. This cross-disciplinary area has been surveyed by a number of papers from graphics~\cite{smelik2014survey, Kim2018, vanegas2010modelling}, architecture~\cite{Spencer2019}, and physics~\cite{Barthelemy2011}.
\newline\newline
\cpara{Procedural Modeling.} 
M{\"u}ller et al.~\cite{Parish:2001:PMC} introduced a novel self-sensitive L-System~\cite{prusinkiewicz1986graphical} which was well suited to the construction of a variety of different street patterns. It has also been adapted by follow up work~\cite{benevs2011guided,kelly2007citygen}. Other approaches include simple grids~\cite{greuter2003real}, mesh-based deformable templates~\cite{peng2014computing} and hierarchical domain splitting~\cite{yang2013urban}.
%
%
Outside of cities, the interaction of street networks with terrain and natural features (e.g., rivers and lakes) are important to generation~\cite{emilien2012procedural,Galin2010Procedural}.
One may also apply optimization and simulation in other ways, such as agent-based generation~\cite{song2019townsim}, or quality of life factors (e.g., daylight or distance to the nearest park)~\cite{vanegas2012inverse}.


\cpara{Simulation.} Street networks are shaped by complex sociological, economic, and technological aspects throughout history. A large body of work studies land use modeling~\cite{lechner2006procedural} and other work aims to create more accurate street- and city-scapes by simulation over time, in both contemporary~\cite{weber2009interactive} or historically~\cite{mas2020simulating} contexts. Given two historical city maps, Krecklau et al. create a smooth animation ~\cite{krecklau2012procedural} showing city growth over time, while Vanegas et al. create realistic satellite images form such a prediction~\cite{vanegas2009visualization}. Another approach is to allow the user to author a timeline~\cite{williams2017time}, which is used to simulate the history of the urban area.

\cpara{Interactive Techniques.} Direct authoring of street network graphs, placing junctions (nodes) and streets (edges) is accurate, but highly time consuming. At the other extreme, geospatial databases~\cite{sun2002template} are accurate and fast, but cannot synthesize novel areas. In between these extremes, interactive authoring of street networks often focuses on higher level primitives, such as tensor fields ~\cite{chen2008interactive}, brushing elements onto existing street networks~\cite{emilien2015worldbrush}, or street density and patterns~\cite{galin2011authoring}. The recombination of patches of street networks has been investigated using interactive tools such as expand, scale, replace~\cite{aliaga2008interactive}, blending, warping, growing~\cite{nishida2016example}, or shrinking street network patches~\cite{pueyo2020shrinking}. 
~\vspace{-0.3cm}
\subsection{Generative Models}
\cpara{VAEs.} The standard VAE \cite{KingmaWellingVAE} is an encoder-decoder framework, where the data is encoded into latent variables, and decoded from those latents. Further refinements include more expressive priors \cite{TomczakVampPrior, KingmaIAF, chen2016variational}, or better architectures \cite{ArashNVAE, Maale2019BIVAAV, Ranganath2016HierarchicalVM, Snderby2016LadderVA} which are often hierarchical.
VQVAE~\cite{OordVAE,OordVAE2} introduced vector quantization~\cite{GrayVQ} to generative models. During training, the latents from the encoder are replaced by their closest element from a learned dictionary. A discrete dictionary works well in conjunction with auto-regressive models based on transformers.

\cpara{Transformers.} The transformer architecture \cite{VaswaniAttention} originally arose in the field of Natural Language Processing but has since been adapted in fields as diverse as object detection \cite{CarionDETR, beal2020toward}, image classification \cite{dosovitskiy2021an}, image generation \cite{parmar2018image, esser2020taming} and layout generation \cite{para2020generative, wang2020sceneformer}. A transformer is composed of multiple layers of attention blocks \cite{parikh-etal-2016-decomposable, lin2017structured} which model attention over different positions in the sequence. 
Transformers used as generative models are \emph{Auto-Regressive Models} \cite{OordPixelCNN, esser2020taming} where the generation process occurs step-by-step and the current element depends only on the previously generated elements.
\vspace{-0.2cm}
\subsection{Graph Synthesis Models}
\cpara{Modeling Graph Connectivity.} Deep-generative models for graph-structured data were originally proposed in \cite{kusner2016GVAE}. This was followed by a follow-up work \cite{dai2018syntaxdirected, Guimaraes2017ObjectiveReinforcedGA, li2018learning, you2018graphrnn, de2018molgan, simonovsky2018graphvae} which showed impressive results for molecule generation. Recent methods focus on attaining better performance by using improved architectures \cite{liao2019gran, li2021graph}, handling larger graphs by exploiting sparsity \cite{pmlr-v119-dai20b}, or improving priors \cite{Shi*2020GraphAF, moflow, honda2020graph} with flows. Single step generation with VAEs/GANs \cite{de2018molgan, simonovsky2018graphvae}, while faster than auto-regressive models \cite{Grover2019GraphiteIG, you2018graphrnn, liao2019gran}, tends to be of lower quality.
These methods are used in domains where nodes do not carry any spatial information and mostly focus on generating valid topology. In the street domain, generating both valid geometry and valid topology is required. 

\cpara{Graphs With Geometric Data.} Neural Turtle Graphics \cite{chu2019ntg} treats a road network as a graph where the nodes have 2D coordinates. The model is an encoder-decoder architecture based on GRU \cite{cho-etal-2014-learning} cells in a similar way to Recursive Neural Networks (RNN). For each node, the encoder encodes all the paths which terminate at the node into a hidden state. The decoder works recursively to decode the hidden state into a variable length sequence containing coordinates of the current node, and coordinates of a variable number of neighbouring nodes.
PolyGen \cite{pmlr-v119-nash20a} is a method for generating 3D meshes. PolyGen tackles the graph generation problem by factorizing the graph into two models. First a \textit{Vertex Model} learns the distribution of the nodes/vertices. Second a \textit{Face Model} learns the  connectivity distribution of edges/faces conditioned on the nodes. This enables the modeling of both spatial and topological information. HDMapGen~\cite{mi2021hdmapgen} also employs auto-regressive modeling for smaller graphs.

We observed that methods like PolyGen \cite{pmlr-v119-nash20a} tend to have problems to provide
convincing results when relying on a meaningful topology. We particularly noticed more local errors in connectivity that are ultimately a bit more noticeable than the errors from image-based representations like a distance field. We therefore leverage a robust CNN-based decoder to obtain an image-based representation, followed by a thinning approach and road-tracing method proposed in \cite{roadtracer}.

Alternatively, GANs could also be used for road modeling~\cite{hartmann2017streetgan,owaki2020roadnetgan}, but it is difficult to compete with auto-regressive models.


\section{Methodology} 
\subsection{Overview}
Our system consists of multiple parts, an overview of which are introduced in Fig.~\ref{fig:overview}. A graph is traditionally described as $G(V, E)$ where $V$ are the vertices and $E$ the edges; alternatively, graphs can be represented in the image domain by drawing the graph as a 2D image $\mathbf{I} \in \mathbb{R}^{H \times W \times C}$, with $H$ and $W$ being the height and width in pixels respectively and $C$ being the number of color channels, or a distance field of the same resolution. We first learn a compressed representation of the road layout in the image domain by training a single-level VQVAE called the \cmod. This gives us an Index Field, $\mathbf{z}_q \in \mathbb{Z^+}^{H_q \times W_q}$, which contains indices into a learned dictionary of latents.  We experimented with various different image-based representations of a street graph and identified distance fields as giving the best results.



The compressed Index Field representation reduces the sequence length for the decoders which operates on the latent representation. This compression is achieved spatially since $H_q << H$ and $W_q << W$, as well as through the re-use of values in the indexed dictionary. The sliding window generation scheme further limits the effective sequence length, but it requires additional supervision, e.g., in the form of high priority roads, a land water map, and a density map to achieve global consistency. Combining these approaches allows us to model road networks corresponding to over $22 km \times 22 km$ in the real world, which is over two orders of magnitude larger than prior art. Further, we can achieve this with a transformer, which is otherwise highly sensitive to long sequence lengths.

\vspace{-0.2cm}
\subsection{Data Preparation}
The input to our training procedure are square \emph{crops} of city areas with each crop being partitioned into a regular grid of equally sized \emph{cells}. The reasoning behind this partitioning is to have one cell per index to conform to the compression factor of the \cmod. 

\cpara{Zoom Level and Projection.}
We use the Mercator projection for our map data and could identify a crop width of $1223m$ on the equator (zoom level $z=15$ following OpenStreetMap standards) to be a good trade-off between being large enough to show enough spatial context and at the same time being small enough to cause feasible sequence lengths per cell for our transformer models.


\cpara{Partitioning Into Cells.}
Each crop is partitioned into a regular grid of $H_q \times W_q$ equally sized cells to conform to the compression factor of the \cmod. In our experiments, we usually choose an input resolution to the \cmod of $256 \times 256$ pixels and a compression ratio to obtain $16 \times 16$ indices as output. Analogously, we subdivide each crop into $16 \times 16$ cells which results in an effective cell extent of $76.44m \times 76.44m$. 

\cpara{Density Maps.} Our generation is conditioned on a locally computed street density function (street segment length per unit area). The density for each cell is computed at a resolution of $4\times 4$, with each sample's center over an extent of $1223m \times 1223m$. Computing this map is expensive during test and training time; to this end we accelerate it by pre-computing a global map of densities at a resolution of $76.44m / 4 = 19.11m$. This map may be efficiently convolved for any given sample within a cell to compute the local density.


\cpara{Street Types.}
We observed unsatisfying results in early experiments when trying to synthesize a whole street network in one go without distinguishing between the street types (e.g., motorway, residential road); therefore, our goal is to generate a street network in a hierarchical manner by starting with high-priority streets such as highways and to condition the generation of lower-priority streets on already generated streets of higher priority. To this end, we define two priority classes $P_1$ and $P_2$, with $P_1$ being streets of high priority (e.g., motorway, trunk) and $P_2$ being streets of low priority (e.g., residential road, living street), and split the data accordingly. For the proof of concept proposed in this paper, we focus on the generation of $P_2$ streets and assume an already existing network of $P_1$ streets. Please refer to Appendix \ref{appendix:street_types} for a complete assignment list of street types to priority classes.




\begin{figure}[t!]
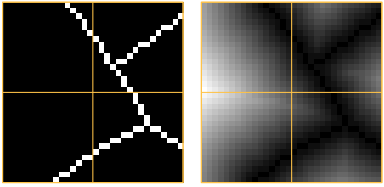

    \centering
    \begin{subfigure}[c]{0.3\linewidth}
        \input{figures/discretization/fig-bresenham.tikz}
    \end{subfigure}
    ~
    \begin{subfigure}[c]{0.3\linewidth}
        \input{figures/discretization/fig-distance_field.tikz}
    \end{subfigure}
    \caption{Two different image-based representations of an example crop partitioned into $2 \times 2$ cells \protect\tikz{\protect\draw[our-orange] (0,0) rectangle (5pt,5pt);}. Left: The crop in the image domain ($32 \times 32$ px, $16 \times 16$ px per cell) rendered using Bresenham line rasterizer. Right: Distance Field.}
    \label{fig:discretiz}
\vspace{-0.5cm}
\end{figure}

\subsection{\cmod}
We leverage the VQGAN implementation of Esser et al.~\cite{esser2020taming} to learn a dictionary of context-rich latent codes. However, we could not observe any improvement when using the discriminator $D$ but only use a combination of reconstruction loss, perceptual loss and dictionary loss. Due to this missing adversarial component, we consider our \cmod to be a VQVAE. We experimented with a two image-based representations of a street network to be used as input to our model -- \emph{simple graph drawings} and \emph{distance fields} (See Fig.~\ref{fig:discretiz}):

\cpara{Simple Graph Drawings.}
To represent a graph as a simple image with uniform spatial properties, we draw each edge using a line rasterizer. We use the Bresenham algorithm~\cite{bresenham} due to its simplicity. However, we observe problems with thin one-pixel-wide edges used as input to the VQVAE -- the reconstructions tend to be fragmented and do not retain the important topology of the underlying graph. We therefore experimented with morphological dilation to obtain a better reconstruction quality. This however did not properly solve the mentioned issues. We hypothesize that the VQVAE needs a smooth representation to process information about the street network not only at pixels corresponding to actual elements of the graph, but at each pixel of the input image. This is a condition that a thin rendering does not possess as it has sharp transitions (e.g., between road and non-road regions).

\cpara{Distance Fields.}
We use distance fields $\mathcal{D}$ as a smooth image-based representation of the graph. While it would be possible to compute the fields analytically on-demand, we found this to be a performance bottleneck when training. Therefore, we first compute the rasterized graph as a simple graph drawing (as above) and then pre-compute the distance field, interpreting all pixels corresponding to the graph to have a distance of $0$. While this approximation reduces the accuracy of the distance fields, we could not observe a significant drop of the output quality and therefore rasterize all image-based street network data prior to the distance field computation.

\subsection{\imod}
Once we have a trained \cmod, we use the model to compress crops from renders of city areas into an index field. We compress into a field having a spatial resolution that is per axis $16$ times smaller than the input resolution. We then train a transformer on this field following \cite{esser2020taming}. Our index field is ordered into a sequence $s$. The transformer is trained with the next-token prediction task - given the partial sequence $s_{<i}$, predict the next token. This leads to the following probability model $p(s) = \prod_i p(s_i | s_{<i})$. Extending the model to generate probabilities, conditioned on a context $c$ (e.g., street density or a land-water map) leads to the following model:
\begin{equation}
    p(s) = \prod_i p(s_i | s_{<i} , c)
    \label{Eq:transformer_main}
\end{equation}

\cpara{Architecture and Training.}
The architecture of the transformer is a standard encoder-decoder model, with bidirectional attention on the encoder, and causal attention on the decoder. We use a sliding-window representing the receptive field of the transformer of size $16 \times 16$ resulting in a sequence length of $256$. We use a learned \textit{positional embedding} where the position assigned to each index is the absolute location within the window. We also need to be able to condition on the provided context. There are two forms of context. The first is the \textit{cell-based context} ($c_c$), which is the quantized representation of the density. The density map is $\frac{1}{16}$th the size of the desired spatial resolution of the graph, while $P_1$ and the land-water map have the same resolution. The cell-based context is encoded by using a linear layer on the scalar value of the cell to project to the embedding dimension of the transformer. The second type of context is the \textit{pixel-based context} ($c_p$) which is the image based representation of given distance fields (e.g., $P_1$ streets) and the land-water map. We extract the patches corresponding to each of the cells in the sliding window from the image representations. The image representations are flattened and projected by a linear layer to the embedding dimension. This is done \textit{independently} for every cell. The two embedding vectors per-cell are concatenated in the channel dimension, and another linear layer projects them again to the embedding dimension. The encoder input sequence is then the concatenation of all the embedding vectors, this time along the sequence dimension. The complete context $c$, in Eq.~\ref{Eq:transformer_main} is the concatenation of the the two contexts on a per-cell basis $c = [c_c; c_p]$.  A transformer encoder encodes this context.
A transformer decoder performs cross-attention into the output of the encoder to create a distribution over the possible tokens and generate $s_i$ as in Eq.~\ref{Eq:transformer_main}.



\cpara{Inference.}
In order to generate an index field, we need a set of conditioning maps - a distance field representing $P_1$, a density map defining the desired street network density per cell and the land-water map. Inference is almost the same as training, where we take the corresponding crops from the appropriate conditioning maps, move the sliding window by 1 cell during the generation process and denote the generated index field as $\mathcal{F}$. Using the sliding window approach, we are able to generate index fields of arbitrary size. For a street network corresponding to an area of about 20km x 20km, we choose both the height $H_{\mathcal{F}}$ and width $W_{\mathcal{F}}$ of the index field to be 256. 


    






\subsection{Decoder} 
We decode the generated index field $\mathcal{F}$ back into a distance field using the trained decoder of the VQVAE. We post-process the distance field to a graph representation. To do this we first apply a threshold, then using a thinning and road-tracing~\cite{roadtracer}, before finally filtering small disconnected parts. This results in a vector street graph output. 

\section{Results}
We trained on the largest 50 US cities by population to extract a large variety of different city crops. Our results show interesting layouts containing many different patterns.

\cpara{Implementation Details.}
Our models are trained on a single node with one to four A100 Nvidia GPUs. For the VQVAE, we use the official VQGAN implementation of \cite{esser2020taming} without a discriminator and the fast transformers library of \cite{katharopoulos_et_al_2020,vyas_et_al_2020} for the \imod. For the results shown in this paper, we choose a learning rate of $3.0e-4$, $8$ attention heads, and usually train for $65k$ gradient updates. We found a warm-up of $1k$ gradient updates to be beneficial and use a cosine annealing afterwards. Training takes about $24$ hours for the VQVAE and $48$ hours for the \imod. For easier data acquisition, we set up our own OpenStreetMap server consisting of a tile server for the land/water maps and leverage the Overpass API to query the vector data of street networks.

\begin{figure*}[h]
    \centering
    \includegraphics[width=0.95\textwidth]{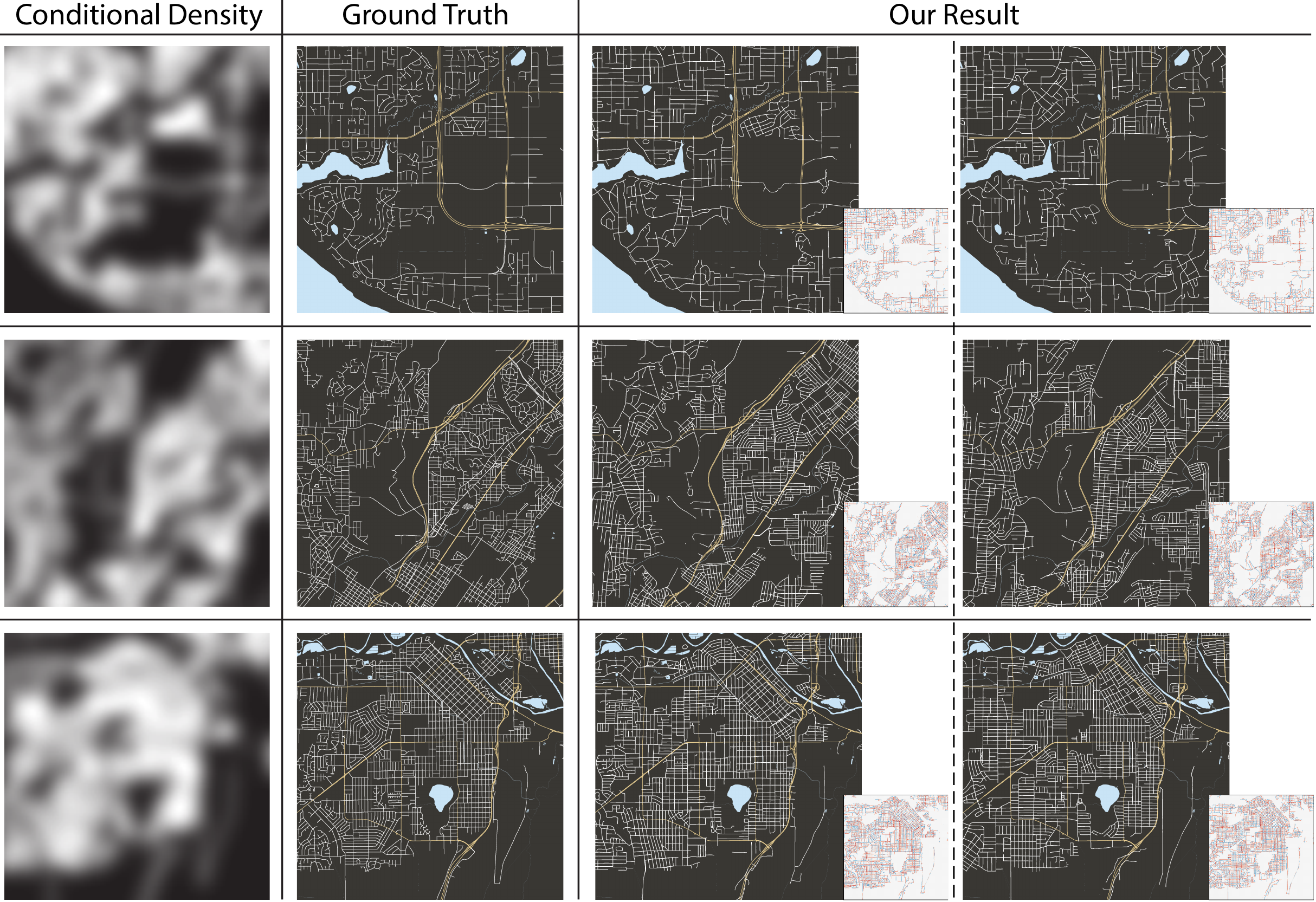}
    \caption{Given the maps for conditioning (land/water, density, existing higher-priority streets), our model is able to generate a large diversity of different street networks. The insets show the difference between the Ground Truth and the generated networks.}
    \label{fig:output_diversity}
\end{figure*}

\cpara{Output Diversity.}
We use our model repeatedly using the same set of condition maps to show the diversity of the generated results. In Fig.~\ref{fig:output_diversity}, we show two different layouts for crops of size $9.8$ km $\times$ $9.8$ km for three US cities. Note how the model adheres to the given conditions (land/water, higher priority streets, density) to create a different layout in each case.

\cpara{Large-Scale Results.}
Using our model in a sliding window fashion we are able to create a city area of arbitrary size. The only limiting factor in this regard is the autoregressive generation which requires the output to be generated sequentially. We validate our system by choosing an area consisting of $256 \times 256$ cells, each cell corresponding to an area of about $76$ m $\times$ $76$ m, the output extent thereby corresponding to about $19.5$ km $\times$ $19.5$ km in the Mercator-projected space. Although our system can be used together with a user-defined density map, we use the ground-truth density map for our results to allow a direct comparison with existing cities. In Fig.~\ref{fig:teaser}, we show an alternative city layout for New York City. Our model is able to adhere to the provided condition maps in a meaningful way and to generate a diverse layout that would be difficult to achieve with state-of-the-art procedural models.

\begin{figure}[b!]
    \centering
    \begin{subfigure}[c]{.45\linewidth}
        \includegraphics[width=0.95\textwidth]{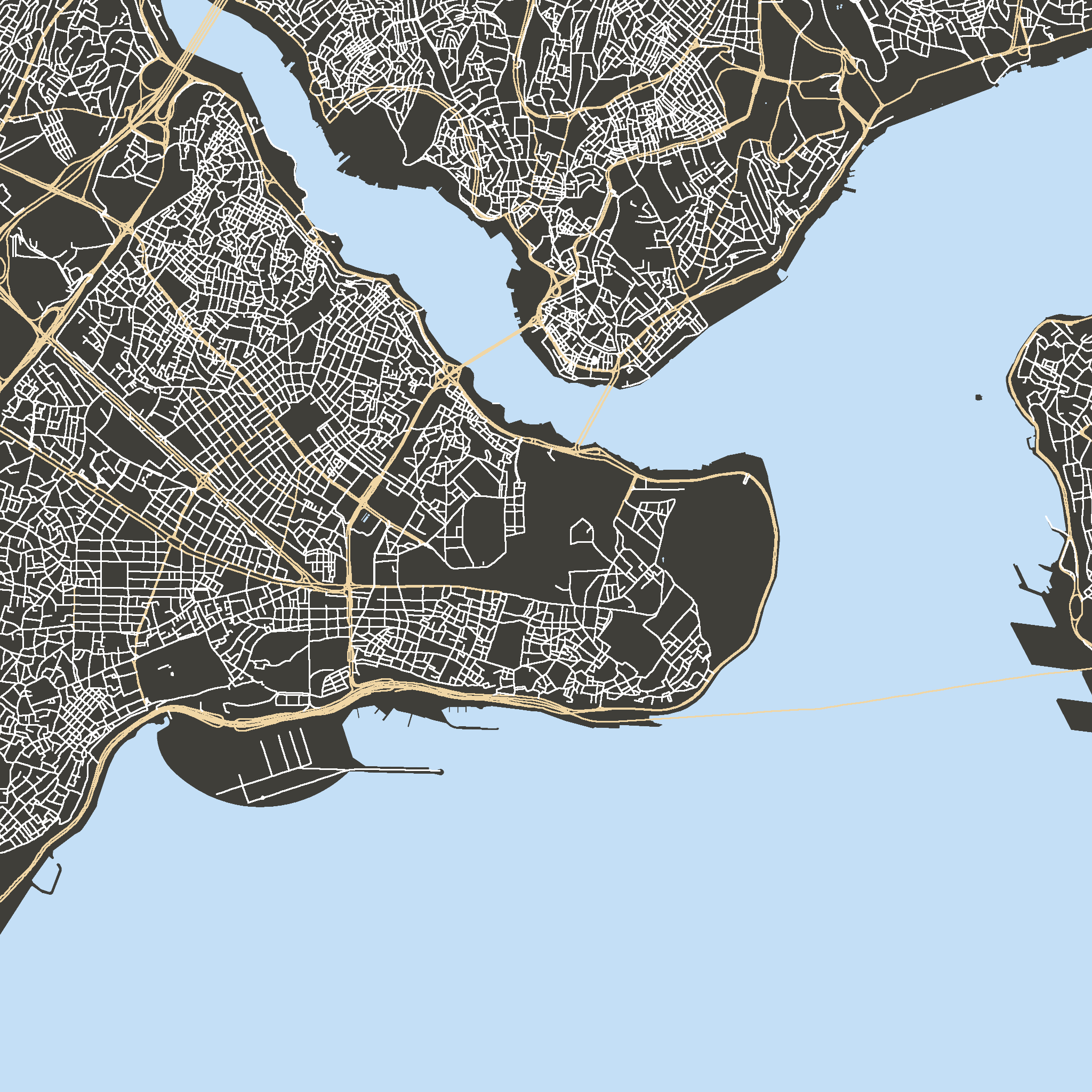}
    \end{subfigure}
    ~
    \begin{subfigure}[c]{.45\linewidth}
        \includegraphics[width=0.95\textwidth]{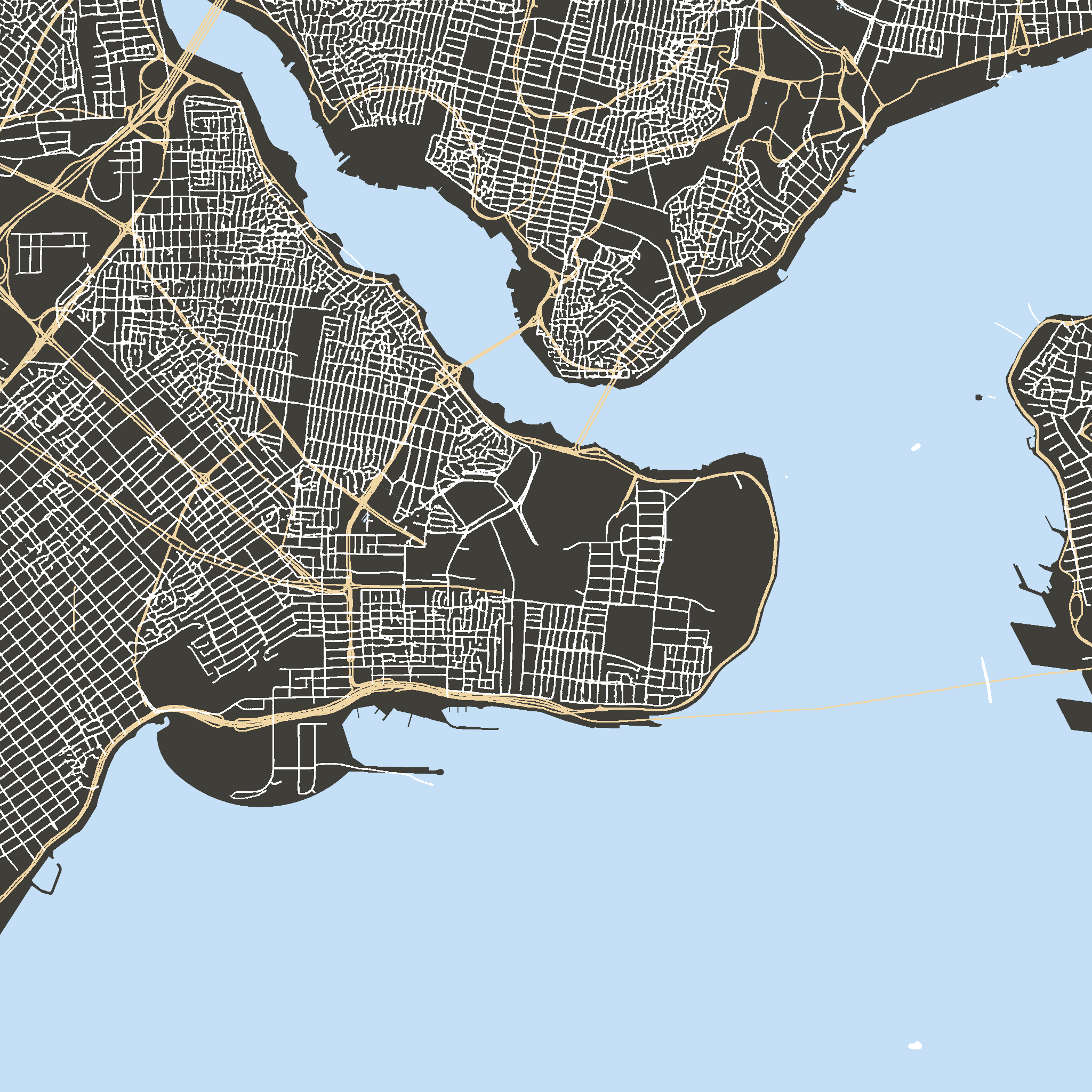}
    \end{subfigure}
    \caption{Our model can be used to do style transfer of street pattern. Left: original street map of Istanbul. Right: regenerated street network with the density map of Istanbul by a network trained only on US street graphs.}
    \label{fig:istanbul}
\end{figure}

\cpara{Style Transfer}
We experimented with using our trained model to transfer the style of streets between different locations. We used our model that was trained on street patterns in US cities to synthesize street networks for cities in other countries. In Fig. \ref{fig:istanbul} we show the application of the model to recreate Istanbul, in an US style, from high-priority streets, density, and coastline.

\begin{figure}[t!]
    \centering
    \begin{subfigure}[c]{.30\linewidth}
        \includegraphics[width=0.95\textwidth]{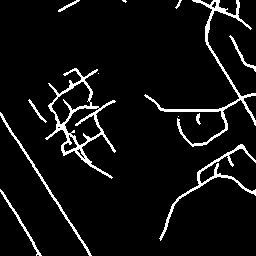}
    \end{subfigure}
    ~
    \begin{subfigure}[c]{.30\linewidth}
        \includegraphics[width=0.95\textwidth]{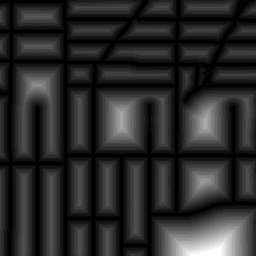}
    \end{subfigure}
    \caption{Left: rendered graphs. Right: distance fields.}
    \label{fig:decoder}
\end{figure}



\cpara{Comparing Statistics to Procedural Models}
We use several statistics to compare our results to procedural cities generated by CityEngine (CE). CE has several parameters for street generation, including the pattern type (grid, organic), average block width, and average block length. We estimate these parameters from real cities and compare our generated street networks to CE results. This comparison mainly illustrates that procedural models do not have enough parameters to recreate the pattern of real street maps. We show an example for Chicago in Fig.~\ref{fig:stats} and more comparisons and an explanation of the statistics in supplementary.

\cpara{Comparison to GANs.}
\begin{figure}[t!]
    \centering
    \begin{subfigure}[c]{.30\linewidth}
        \includegraphics[width=0.95\textwidth]{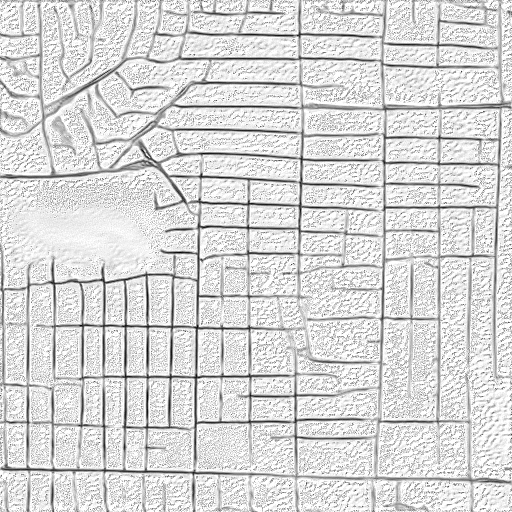}
    \end{subfigure}
    ~
    \begin{subfigure}[c]{.30\linewidth}
        \includegraphics[width=0.95\textwidth]{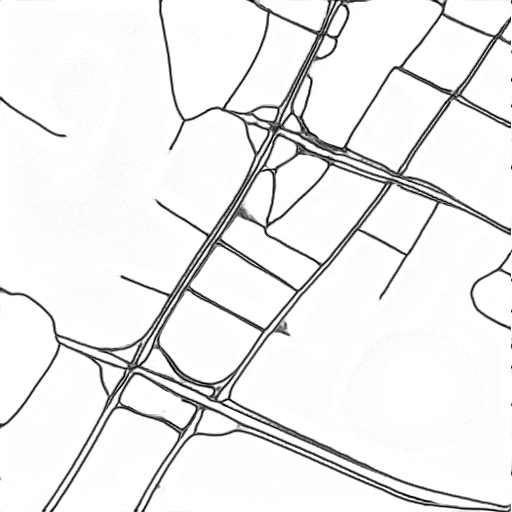}
    \end{subfigure}
    ~
    \begin{subfigure}[c]{.30\linewidth}
        \includegraphics[width=0.95\textwidth]{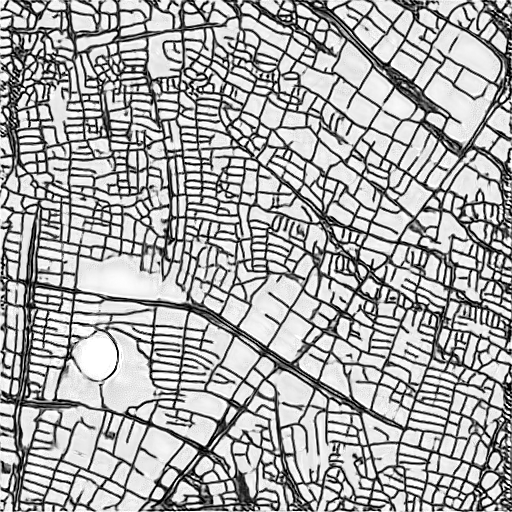}
    \end{subfigure}
    \caption{StyleGAN generated image representations of the street network. We found that the output was too noisy (left), did not have enough sharp features (center), or was too different from the training set (right) to create useful street graphs after thinning.}
    \label{fig:gan}
\end{figure}
Generative Adversarial Networks (GANs) are another competing approach that could be employed for street generation. We show an example in Fig.\ref{fig:gan} generated with StyleGANv2~\cite{Karras2020ada}. While GANs offer the advantage of fast generation times, the output is unusable with the thinning algorithm or other vector graphic extraction algorithms. 

    


\begin{figure}[t!]
    \centering
    \begin{subfigure}[c]{.45\linewidth}
        \includegraphics[width=0.95\textwidth]{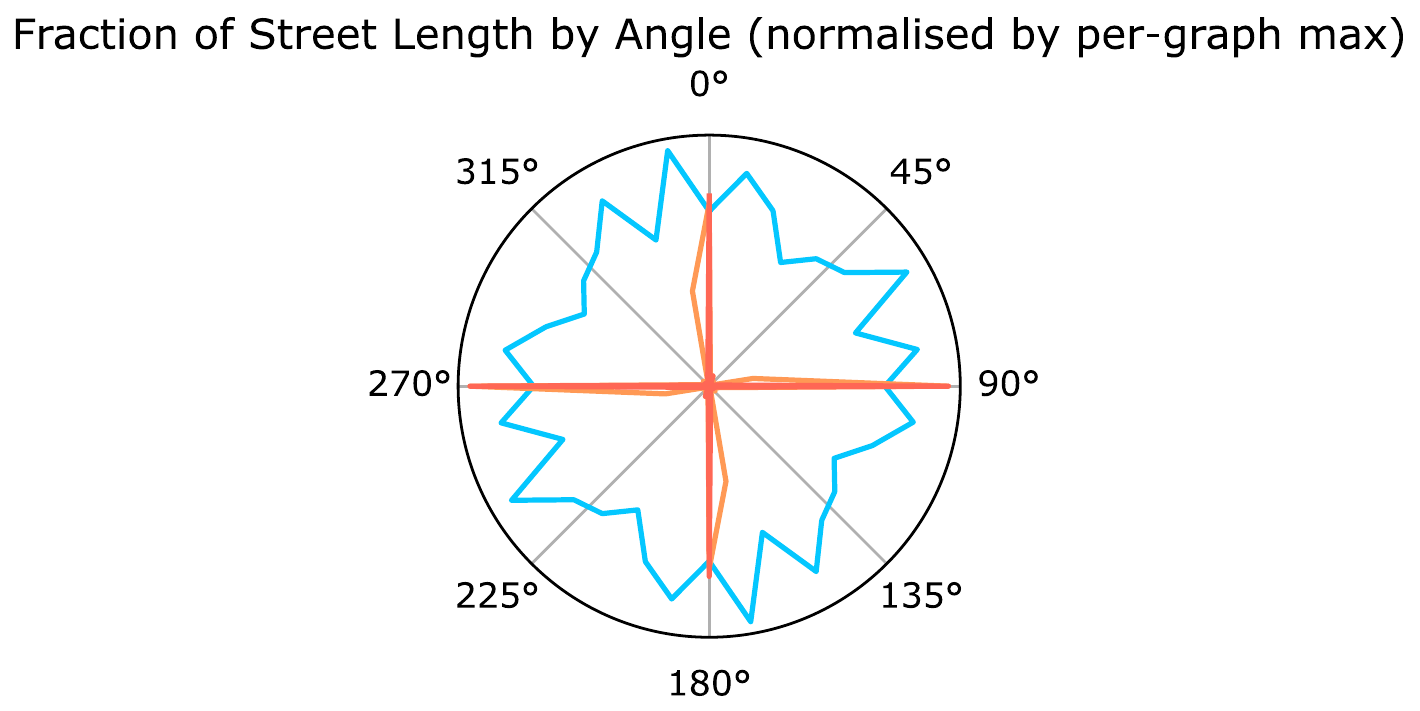}
    \end{subfigure}
    
    \begin{subfigure}[c]{.8\linewidth}
        \includegraphics[width=0.95\textwidth]{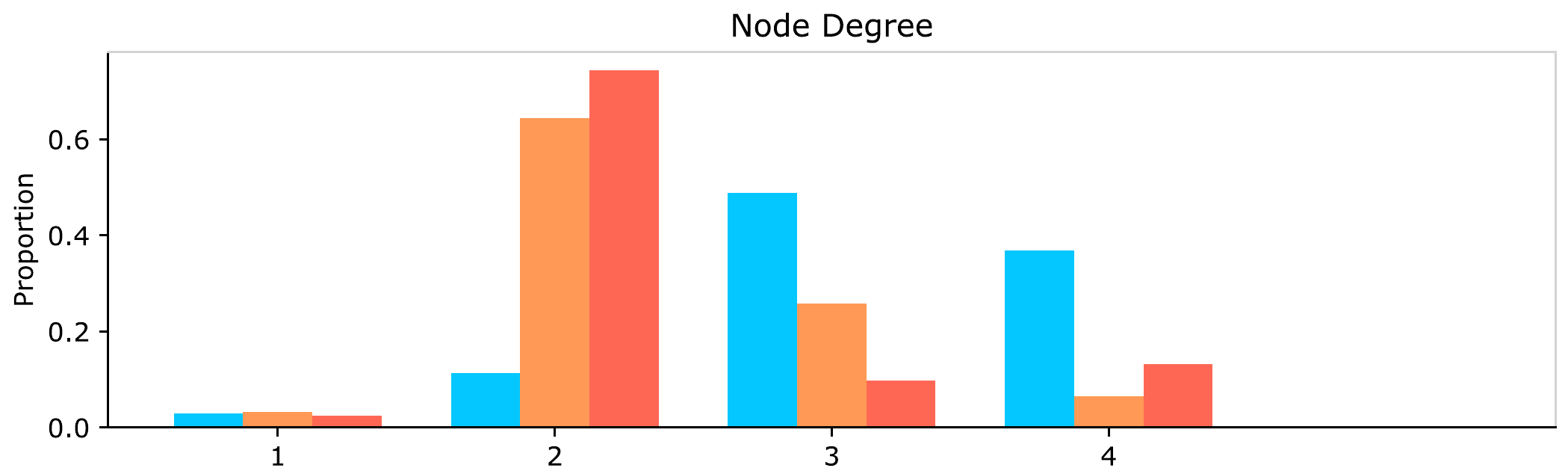}
    \end{subfigure}
    
    \begin{subfigure}[c]{.8\linewidth}
        \includegraphics[width=0.95\textwidth]{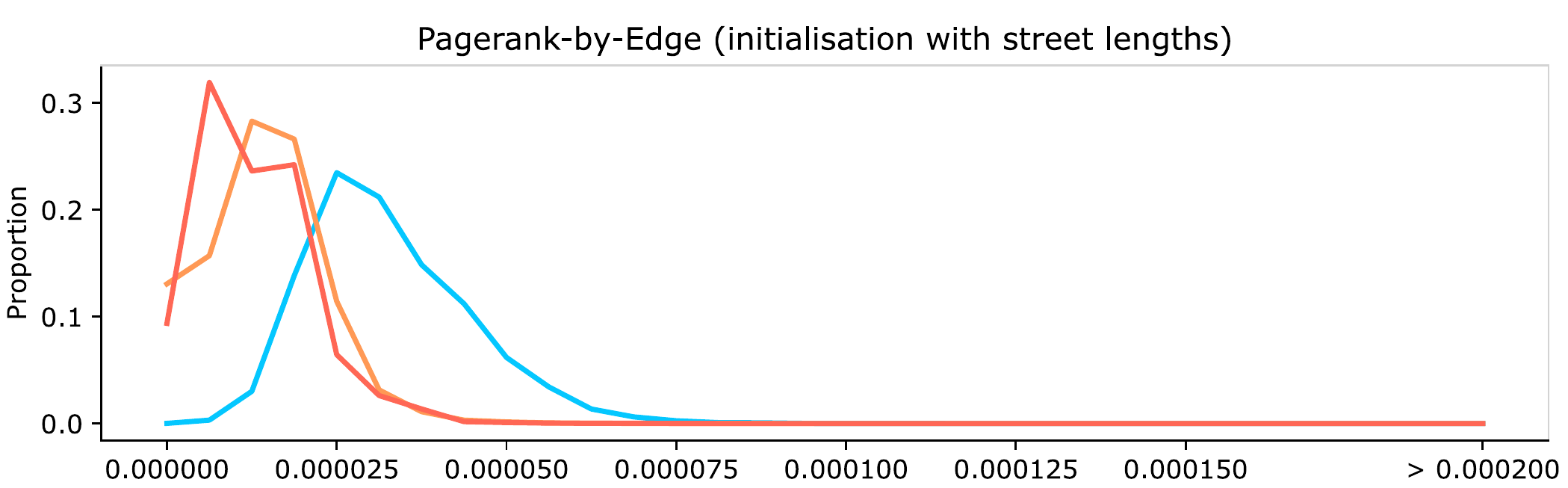}
    \end{subfigure}
    
    \begin{subfigure}[c]{.8\linewidth}
        \includegraphics[width=0.95\textwidth]{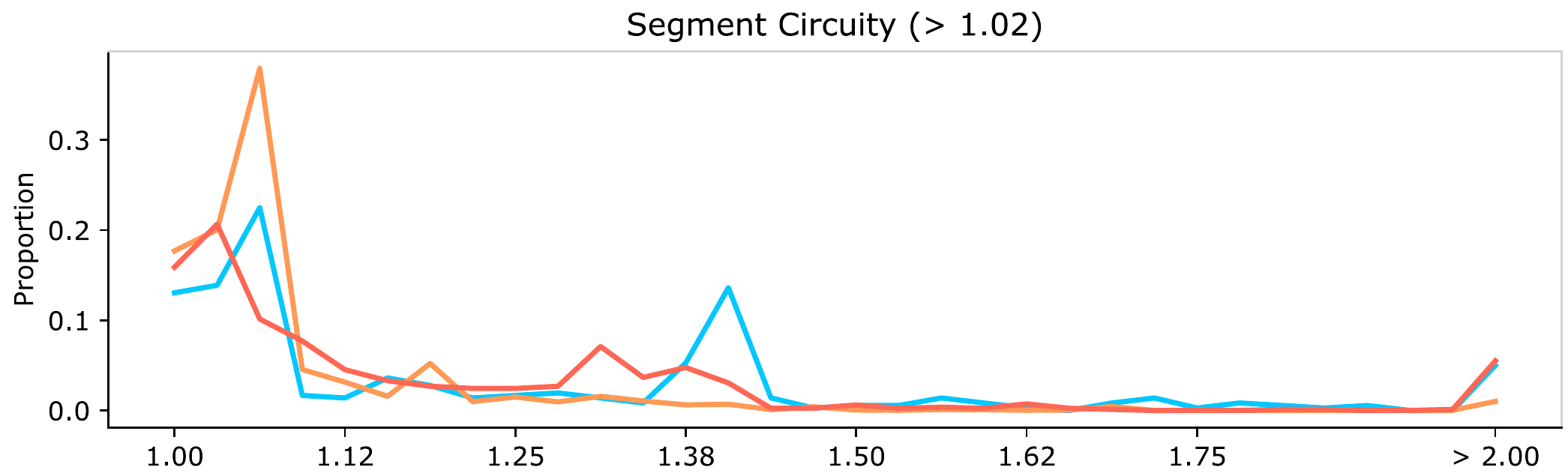}
    \end{subfigure}
    
    \begin{subfigure}[c]{.8\linewidth}
        \includegraphics[width=0.95\textwidth]{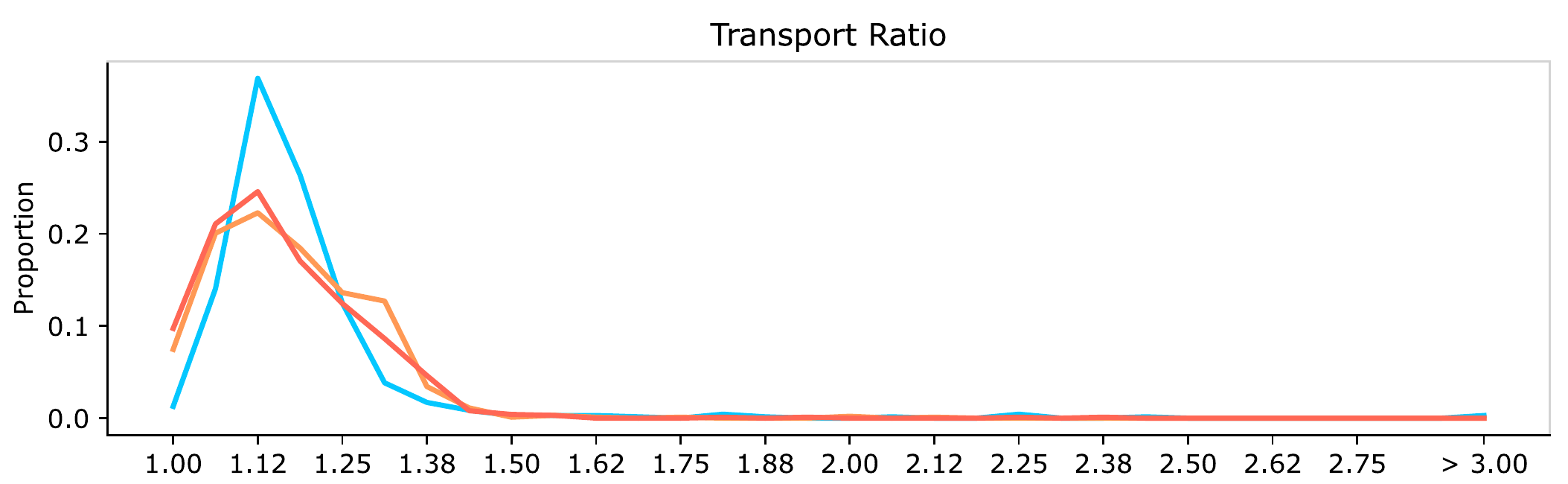}
    \end{subfigure}
    
    \caption{Our model is significantly better able to adhere to statistical numbers of existing cities than the state-of-the art rule-based system CityEngine. Orange: Our model, Red: Ground-Truth, Blue: CityEngine}
    \label{fig:stats}
\end{figure}

\section{Conclusions, Limitations, and Future Work}

We propose a novel framework for modeling large-scale street networks using auto-regressive models. Our results show street networks over two magnitude larger than networks produced by prior art.
Our work has some limitations that could be addressed in future work. For now, conditioning information is manually designed. While this provides end-users with control over the generated layouts, it is still a laborious and time consuming process. It would be desirable to generate conditioning information by other generative models specifying only high-level details. Further, we do not consider terrain as conditioning information in the current iteration of the paper. The main reason was the difficulty of obtaining training data.  In future work, we hope to expand the model to generate parcels and buildings.



\bibliographystyle{ACM-Reference-Format}
\bibliography{bib}


\begin{thebibliography}{95}


\ifx \showCODEN    \undefined \def \showCODEN     #1{\unskip}     \fi
\ifx \showDOI      \undefined \def \showDOI       #1{#1}\fi
\ifx \showISBNx    \undefined \def \showISBNx     #1{\unskip}     \fi
\ifx \showISBNxiii \undefined \def \showISBNxiii  #1{\unskip}     \fi
\ifx \showISSN     \undefined \def \showISSN      #1{\unskip}     \fi
\ifx \showLCCN     \undefined \def \showLCCN      #1{\unskip}     \fi
\ifx \shownote     \undefined \def \shownote      #1{#1}          \fi
\ifx \showarticletitle \undefined \def \showarticletitle #1{#1}   \fi
\ifx \showURL      \undefined \def \showURL       {\relax}        \fi
\providecommand\bibfield[2]{#2}
\providecommand\bibinfo[2]{#2}
\providecommand\natexlab[1]{#1}
\providecommand\showeprint[2][]{arXiv:#2}

\bibitem[Ackley et~al\mbox{.}(1987)]%
        {AckleyBoltzmann}
\bibfield{author}{\bibinfo{person}{David Ackley}, \bibinfo{person}{Geoffrey
  Hinton}, {and} \bibinfo{person}{Terrence Sejnowski}.}
  \bibinfo{year}{1987}\natexlab{}.
\newblock \showarticletitle{A Learning Algorithm for Boltzmann Machines}.
\newblock In \bibinfo{booktitle}{\emph{Readings in Computer Vision}},
  \bibfield{editor}{\bibinfo{person}{Martin~A. Fischler} {and}
  \bibinfo{person}{Oscar Firschein}} (Eds.). \bibinfo{publisher}{Morgan
  Kaufmann}, \bibinfo{address}{San Francisco (CA)}, \bibinfo{pages}{522--533}.
\newblock
\showISBNx{978-0-08-051581-6}
\urldef\tempurl%
\url{https://doi.org/10.1016/B978-0-08-051581-6.50053-2}
\showDOI{\tempurl}


\bibitem[Aliaga et~al\mbox{.}(2008)]%
        {aliaga2008interactive}
\bibfield{author}{\bibinfo{person}{Daniel~G Aliaga},
  \bibinfo{person}{Bed{\v{r}}ich Bene{\v{s}}}, \bibinfo{person}{Carlos~A
  Vanegas}, {and} \bibinfo{person}{Nathan Andrysco}.}
  \bibinfo{year}{2008}\natexlab{}.
\newblock \showarticletitle{Interactive reconfiguration of urban layouts}.
\newblock \bibinfo{journal}{\emph{IEEE Computer Graphics and Applications}}
  \bibinfo{volume}{28}, \bibinfo{number}{3} (\bibinfo{year}{2008}),
  \bibinfo{pages}{38--47}.
\newblock


\bibitem[Barthélemy(2011)]%
        {Barthelemy2011}
\bibfield{author}{\bibinfo{person}{Marc Barthélemy}.}
  \bibinfo{year}{2011}\natexlab{}.
\newblock \showarticletitle{Spatial networks}.
\newblock \bibinfo{journal}{\emph{Physics Reports}} \bibinfo{volume}{499},
  \bibinfo{number}{1} (\bibinfo{year}{2011}), \bibinfo{pages}{1--101}.
\newblock
\showISSN{0370-1573}
\urldef\tempurl%
\url{https://doi.org/10.1016/j.physrep.2010.11.002}
\showDOI{\tempurl}


\bibitem[Bastani et~al\mbox{.}(2018)]%
        {roadtracer}
\bibfield{author}{\bibinfo{person}{F. Bastani}, \bibinfo{person}{S. He},
  \bibinfo{person}{S. Abbar}, \bibinfo{person}{M. Alizadeh},
  \bibinfo{person}{H. Balakrishnan}, \bibinfo{person}{S. Chawla},
  \bibinfo{person}{S. Madden}, {and} \bibinfo{person}{D. DeWitt}.}
  \bibinfo{year}{2018}\natexlab{}.
\newblock \showarticletitle{RoadTracer: Automatic Extraction of Road Networks
  from Aerial Images}. In \bibinfo{booktitle}{\emph{2018 IEEE/CVF Conference on
  Computer Vision and Pattern Recognition (CVPR)}}. \bibinfo{publisher}{IEEE
  Computer Society}, \bibinfo{address}{Los Alamitos, CA, USA},
  \bibinfo{pages}{4720--4728}.
\newblock
\urldef\tempurl%
\url{https://doi.org/10.1109/CVPR.2018.00496}
\showDOI{\tempurl}


\bibitem[Beal et~al\mbox{.}(2020)]%
        {beal2020toward}
\bibfield{author}{\bibinfo{person}{Josh Beal}, \bibinfo{person}{Eric Kim},
  \bibinfo{person}{Eric Tzeng}, \bibinfo{person}{Dong~Huk Park},
  \bibinfo{person}{Andrew Zhai}, {and} \bibinfo{person}{Dmitry Kislyuk}.}
  \bibinfo{year}{2020}\natexlab{}.
\newblock \showarticletitle{Toward Transformer-Based Object Detection}.
\newblock \bibinfo{journal}{\emph{arXiv preprint arXiv:2012.09958}}
  (\bibinfo{year}{2020}).
\newblock


\bibitem[Beltagy et~al\mbox{.}(2020)]%
        {Beltagy2020LongformerTL}
\bibfield{author}{\bibinfo{person}{Iz Beltagy}, \bibinfo{person}{Matthew~E.
  Peters}, {and} \bibinfo{person}{Arman Cohan}.}
  \bibinfo{year}{2020}\natexlab{}.
\newblock \showarticletitle{Longformer: The Long-Document Transformer}.
\newblock \bibinfo{journal}{\emph{ArXiv}}  \bibinfo{volume}{abs/2004.05150}
  (\bibinfo{year}{2020}).
\newblock


\bibitem[Bene{\v{s}} et~al\mbox{.}(2011)]%
        {benevs2011guided}
\bibfield{author}{\bibinfo{person}{Bedrich Bene{\v{s}}},
  \bibinfo{person}{Ondrej {\v{S}}t'ava}, \bibinfo{person}{Radomir M{\v{e}}ch},
  {and} \bibinfo{person}{Gavin Miller}.} \bibinfo{year}{2011}\natexlab{}.
\newblock \showarticletitle{Guided procedural modeling}. In
  \bibinfo{booktitle}{\emph{Computer Graphics Forum}},
  Vol.~\bibinfo{volume}{30}. Wiley Online Library, \bibinfo{pages}{325--334}.
\newblock


\bibitem[Boeing(2019)]%
        {boeing2019morphology}
\bibfield{author}{\bibinfo{person}{Geoff Boeing}.}
  \bibinfo{year}{2019}\natexlab{}.
\newblock \showarticletitle{The morphology and circuity of walkable and
  drivable street networks}.
\newblock In \bibinfo{booktitle}{\emph{The mathematics of urban morphology}}.
  \bibinfo{publisher}{Springer}, \bibinfo{pages}{271--287}.
\newblock


\bibitem[Boeing(2020)]%
        {boeing2020multi}
\bibfield{author}{\bibinfo{person}{Geoff Boeing}.}
  \bibinfo{year}{2020}\natexlab{}.
\newblock \showarticletitle{A multi-scale analysis of 27,000 urban street
  networks: Every US city, town, urbanized area, and Zillow neighborhood}.
\newblock \bibinfo{journal}{\emph{Environment and Planning B: Urban Analytics
  and City Science}} \bibinfo{volume}{47}, \bibinfo{number}{4}
  (\bibinfo{year}{2020}), \bibinfo{pages}{590--608}.
\newblock


\bibitem[Bresenham(1965)]%
        {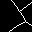}
\bibfield{author}{\bibinfo{person}{J.~E. Bresenham}.}
  \bibinfo{year}{1965}\natexlab{}.
\newblock \showarticletitle{Algorithm for computer control of a digital
  plotter}.
\newblock \bibinfo{journal}{\emph{IBM Systems Journal}} \bibinfo{volume}{4},
  \bibinfo{number}{1} (\bibinfo{year}{1965}), \bibinfo{pages}{25--30}.
\newblock
\urldef\tempurl%
\url{https://doi.org/10.1147/sj.41.0025}
\showDOI{\tempurl}


\bibitem[Carion et~al\mbox{.}(2020)]%
        {CarionDETR}
\bibfield{author}{\bibinfo{person}{Nicolas Carion}, \bibinfo{person}{Francisco
  Massa}, \bibinfo{person}{Gabriel Synnaeve}, \bibinfo{person}{Nicolas
  Usunier}, \bibinfo{person}{Alexander Kirillov}, {and} \bibinfo{person}{Sergey
  Zagoruyko}.} \bibinfo{year}{2020}\natexlab{}.
\newblock \showarticletitle{End-to-End Object Detection with Transformers}. In
  \bibinfo{booktitle}{\emph{Computer Vision - {ECCV} 2020 - 16th European
  Conference, Glasgow, UK, August 23-28, 2020, Proceedings, Part {I}}}
  \emph{(\bibinfo{series}{Lecture Notes in Computer Science},
  Vol.~\bibinfo{volume}{12346})}, \bibfield{editor}{\bibinfo{person}{Andrea
  Vedaldi}, \bibinfo{person}{Horst Bischof}, \bibinfo{person}{Thomas Brox},
  {and} \bibinfo{person}{Jan{-}Michael Frahm}} (Eds.).
  \bibinfo{publisher}{Springer}, \bibinfo{pages}{213--229}.
\newblock
\urldef\tempurl%
\url{https://doi.org/10.1007/978-3-030-58452-8\_13}
\showDOI{\tempurl}


\bibitem[Chen et~al\mbox{.}(2008)]%
        {chen2008interactive}
\bibfield{author}{\bibinfo{person}{Guoning Chen}, \bibinfo{person}{Gregory
  Esch}, \bibinfo{person}{Peter Wonka}, \bibinfo{person}{Pascal M{\"u}ller},
  {and} \bibinfo{person}{Eugene Zhang}.} \bibinfo{year}{2008}\natexlab{}.
\newblock \showarticletitle{Interactive procedural street modeling}.
\newblock \bibinfo{journal}{\emph{ACM Transactions on Graphics (TOG)}}
  \bibinfo{volume}{27}, \bibinfo{number}{3} (\bibinfo{year}{2008}),
  \bibinfo{pages}{1--10}.
\newblock


\bibitem[Chen et~al\mbox{.}(2016)]%
        {chen2016variational}
\bibfield{author}{\bibinfo{person}{Xi Chen}, \bibinfo{person}{Diederik~P
  Kingma}, \bibinfo{person}{Tim Salimans}, \bibinfo{person}{Yan Duan},
  \bibinfo{person}{Prafulla Dhariwal}, \bibinfo{person}{John Schulman},
  \bibinfo{person}{Ilya Sutskever}, {and} \bibinfo{person}{Pieter Abbeel}.}
  \bibinfo{year}{2016}\natexlab{}.
\newblock \showarticletitle{Variational lossy autoencoder}.
\newblock \bibinfo{journal}{\emph{arXiv preprint arXiv:1611.02731}}
  (\bibinfo{year}{2016}).
\newblock


\bibitem[Cho et~al\mbox{.}(2014)]%
        {cho-etal-2014-learning}
\bibfield{author}{\bibinfo{person}{Kyunghyun Cho}, \bibinfo{person}{Bart van
  Merri{\"e}nboer}, \bibinfo{person}{Caglar Gulcehre}, \bibinfo{person}{Dzmitry
  Bahdanau}, \bibinfo{person}{Fethi Bougares}, \bibinfo{person}{Holger
  Schwenk}, {and} \bibinfo{person}{Yoshua Bengio}.}
  \bibinfo{year}{2014}\natexlab{}.
\newblock \showarticletitle{Learning Phrase Representations using {RNN}
  Encoder{--}Decoder for Statistical Machine Translation}. In
  \bibinfo{booktitle}{\emph{Proceedings of the 2014 Conference on Empirical
  Methods in Natural Language Processing ({EMNLP})}}.
  \bibinfo{publisher}{Association for Computational Linguistics},
  \bibinfo{address}{Doha, Qatar}, \bibinfo{pages}{1724--1734}.
\newblock
\urldef\tempurl%
\url{https://doi.org/10.3115/v1/D14-1179}
\showDOI{\tempurl}


\bibitem[Chu et~al\mbox{.}(2019)]%
        {chu2019ntg}
\bibfield{author}{\bibinfo{person}{Hang Chu}, \bibinfo{person}{Daiqing Li},
  \bibinfo{person}{David Acuna}, \bibinfo{person}{Amlan Kar},
  \bibinfo{person}{Maria Shugrina}, \bibinfo{person}{Xinkai Wei},
  \bibinfo{person}{Ming-Yu Liu}, \bibinfo{person}{Antonio Torralba}, {and}
  \bibinfo{person}{Sanja Fidler}.} \bibinfo{year}{2019}\natexlab{}.
\newblock \showarticletitle{Neural Turtle Graphics for Modeling City Road
  Layouts}. In \bibinfo{booktitle}{\emph{ICCV}}.
\newblock


\bibitem[Dai et~al\mbox{.}(2020)]%
        {pmlr-v119-dai20b}
\bibfield{author}{\bibinfo{person}{Hanjun Dai}, \bibinfo{person}{Azade Nazi},
  \bibinfo{person}{Yujia Li}, \bibinfo{person}{Bo Dai}, {and}
  \bibinfo{person}{Dale Schuurmans}.} \bibinfo{year}{2020}\natexlab{}.
\newblock \showarticletitle{Scalable Deep Generative Modeling for Sparse
  Graphs}. In \bibinfo{booktitle}{\emph{Proceedings of the 37th International
  Conference on Machine Learning}} \emph{(\bibinfo{series}{Proceedings of
  Machine Learning Research}, Vol.~\bibinfo{volume}{119})},
  \bibfield{editor}{\bibinfo{person}{Hal~Daumé III} {and}
  \bibinfo{person}{Aarti Singh}} (Eds.). \bibinfo{publisher}{PMLR},
  \bibinfo{pages}{2302--2312}.
\newblock
\urldef\tempurl%
\url{http://proceedings.mlr.press/v119/dai20b.html}
\showURL{%
\tempurl}


\bibitem[Dai et~al\mbox{.}(2018)]%
        {dai2018syntaxdirected}
\bibfield{author}{\bibinfo{person}{Hanjun Dai}, \bibinfo{person}{Yingtao Tian},
  \bibinfo{person}{Bo Dai}, \bibinfo{person}{Steven Skiena}, {and}
  \bibinfo{person}{Le Song}.} \bibinfo{year}{2018}\natexlab{}.
\newblock \showarticletitle{Syntax-Directed Variational Autoencoder for
  Structured Data}. In \bibinfo{booktitle}{\emph{International Conference on
  Learning Representations}}.
\newblock
\urldef\tempurl%
\url{https://openreview.net/forum?id=SyqShMZRb}
\showURL{%
\tempurl}


\bibitem[De~Cao and Kipf(2018)]%
        {de2018molgan}
\bibfield{author}{\bibinfo{person}{Nicola De~Cao} {and} \bibinfo{person}{Thomas
  Kipf}.} \bibinfo{year}{2018}\natexlab{}.
\newblock \showarticletitle{{MolGAN: An implicit generative model for small
  molecular graphs}}.
\newblock \bibinfo{journal}{\emph{ICML 2018 workshop on Theoretical Foundations
  and Applications of Deep Generative Models}} (\bibinfo{year}{2018}).
\newblock


\bibitem[Dosovitskiy et~al\mbox{.}(2021)]%
        {dosovitskiy2021an}
\bibfield{author}{\bibinfo{person}{Alexey Dosovitskiy}, \bibinfo{person}{Lucas
  Beyer}, \bibinfo{person}{Alexander Kolesnikov}, \bibinfo{person}{Dirk
  Weissenborn}, \bibinfo{person}{Xiaohua Zhai}, \bibinfo{person}{Thomas
  Unterthiner}, \bibinfo{person}{Mostafa Dehghani}, \bibinfo{person}{Matthias
  Minderer}, \bibinfo{person}{Georg Heigold}, \bibinfo{person}{Sylvain Gelly},
  \bibinfo{person}{Jakob Uszkoreit}, {and} \bibinfo{person}{Neil Houlsby}.}
  \bibinfo{year}{2021}\natexlab{}.
\newblock \showarticletitle{An Image is Worth 16x16 Words: Transformers for
  Image Recognition at Scale}. In \bibinfo{booktitle}{\emph{International
  Conference on Learning Representations}}.
\newblock
\urldef\tempurl%
\url{https://openreview.net/forum?id=YicbFdNTTy}
\showURL{%
\tempurl}


\bibitem[Emilien et~al\mbox{.}(2012)]%
        {emilien2012procedural}
\bibfield{author}{\bibinfo{person}{Arnaud Emilien}, \bibinfo{person}{Adrien
  Bernhardt}, \bibinfo{person}{Adrien Peytavie}, \bibinfo{person}{Marie-Paule
  Cani}, {and} \bibinfo{person}{Eric Galin}.} \bibinfo{year}{2012}\natexlab{}.
\newblock \showarticletitle{Procedural generation of villages on arbitrary
  terrains}.
\newblock \bibinfo{journal}{\emph{The Visual Computer}} \bibinfo{volume}{28},
  \bibinfo{number}{6} (\bibinfo{year}{2012}), \bibinfo{pages}{809--818}.
\newblock


\bibitem[Emilien et~al\mbox{.}(2015)]%
        {emilien2015worldbrush}
\bibfield{author}{\bibinfo{person}{Arnaud Emilien}, \bibinfo{person}{Ulysse
  Vimont}, \bibinfo{person}{Marie-Paule Cani}, \bibinfo{person}{Pierre Poulin},
  {and} \bibinfo{person}{Bedrich Benes}.} \bibinfo{year}{2015}\natexlab{}.
\newblock \showarticletitle{Worldbrush: Interactive example-based synthesis of
  procedural virtual worlds}.
\newblock \bibinfo{journal}{\emph{ACM Transactions on Graphics (TOG)}}
  \bibinfo{volume}{34}, \bibinfo{number}{4} (\bibinfo{year}{2015}),
  \bibinfo{pages}{1--11}.
\newblock


\bibitem[Esser et~al\mbox{.}(2020)]%
        {esser2020taming}
\bibfield{author}{\bibinfo{person}{Patrick Esser}, \bibinfo{person}{Robin
  Rombach}, {and} \bibinfo{person}{Björn Ommer}.}
  \bibinfo{year}{2020}\natexlab{}.
\newblock \bibinfo{title}{Taming Transformers for High-Resolution Image
  Synthesis}.
\newblock
\newblock
\showeprint[arxiv]{2012.09841}~[cs.CV]


\bibitem[Galin et~al\mbox{.}(2011)]%
        {galin2011authoring}
\bibfield{author}{\bibinfo{person}{Eric Galin}, \bibinfo{person}{Adrien
  Peytavie}, \bibinfo{person}{Eric Gu{\'e}rin}, {and}
  \bibinfo{person}{Bed{\v{r}}ich Bene{\v{s}}}.}
  \bibinfo{year}{2011}\natexlab{}.
\newblock \showarticletitle{Authoring hierarchical road networks}. In
  \bibinfo{booktitle}{\emph{Computer Graphics Forum}},
  Vol.~\bibinfo{volume}{30}. Wiley Online Library, \bibinfo{pages}{2021--2030}.
\newblock


\bibitem[Galin et~al\mbox{.}(2010)]%
        {Galin2010Procedural}
\bibfield{author}{\bibinfo{person}{E. Galin}, \bibinfo{person}{A. Peytavie},
  \bibinfo{person}{N. Maréchal}, {and} \bibinfo{person}{E. Guérin}.}
  \bibinfo{year}{2010}\natexlab{}.
\newblock \showarticletitle{Procedural Generation of Roads}.
\newblock \bibinfo{journal}{\emph{Computer Graphics Forum}}
  \bibinfo{volume}{29}, \bibinfo{number}{2} (\bibinfo{year}{2010}),
  \bibinfo{pages}{429--438}.
\newblock
\urldef\tempurl%
\url{https://doi.org/10.1111/j.1467-8659.2009.01612.x}
\showDOI{\tempurl}
\showeprint{https://onlinelibrary.wiley.com/doi/pdf/10.1111/j.1467-8659.2009.01612.x}


\bibitem[Gersho and Gray(1991)]%
        {GrayVQ}
\bibfield{author}{\bibinfo{person}{Allen Gersho} {and}
  \bibinfo{person}{Robert~M. Gray}.} \bibinfo{year}{1991}\natexlab{}.
\newblock \bibinfo{booktitle}{\emph{Vector Quantization and Signal
  Compression}}.
\newblock \bibinfo{publisher}{Kluwer Academic Publishers},
  \bibinfo{address}{USA}.
\newblock
\showISBNx{0792391810}


\bibitem[Ghosh et~al\mbox{.}(2020)]%
        {Ghosh2020From}
\bibfield{author}{\bibinfo{person}{Partha Ghosh}, \bibinfo{person}{Mehdi S.~M.
  Sajjadi}, \bibinfo{person}{Antonio Vergari}, \bibinfo{person}{Michael Black},
  {and} \bibinfo{person}{Bernhard Scholkopf}.} \bibinfo{year}{2020}\natexlab{}.
\newblock \showarticletitle{From Variational to Deterministic Autoencoders}. In
  \bibinfo{booktitle}{\emph{International Conference on Learning
  Representations}}.
\newblock
\urldef\tempurl%
\url{https://openreview.net/forum?id=S1g7tpEYDS}
\showURL{%
\tempurl}


\bibitem[Goodfellow et~al\mbox{.}(2014)]%
        {GoodfellowGAN}
\bibfield{author}{\bibinfo{person}{Ian Goodfellow}, \bibinfo{person}{Jean
  Pouget-Abadie}, \bibinfo{person}{Mehdi Mirza}, \bibinfo{person}{Bing Xu},
  \bibinfo{person}{David Warde-Farley}, \bibinfo{person}{Sherjil Ozair},
  \bibinfo{person}{Aaron Courville}, {and} \bibinfo{person}{Yoshua Bengio}.}
  \bibinfo{year}{2014}\natexlab{}.
\newblock \showarticletitle{Generative Adversarial Nets}. In
  \bibinfo{booktitle}{\emph{Advances in Neural Information Processing
  Systems}}, \bibfield{editor}{\bibinfo{person}{Z.~Ghahramani},
  \bibinfo{person}{M.~Welling}, \bibinfo{person}{C.~Cortes},
  \bibinfo{person}{N.~Lawrence}, {and} \bibinfo{person}{K.~Q. Weinberger}}
  (Eds.), Vol.~\bibinfo{volume}{27}. \bibinfo{publisher}{Curran Associates,
  Inc.}
\newblock
\urldef\tempurl%
\url{https://proceedings.neurips.cc/paper/2014/file/5ca3e9b122f61f8f06494c97b1afccf3-Paper.pdf}
\showURL{%
\tempurl}


\bibitem[Greuter et~al\mbox{.}(2003)]%
        {greuter2003real}
\bibfield{author}{\bibinfo{person}{Stefan Greuter}, \bibinfo{person}{Jeremy
  Parker}, \bibinfo{person}{Nigel Stewart}, {and} \bibinfo{person}{Geoff
  Leach}.} \bibinfo{year}{2003}\natexlab{}.
\newblock \showarticletitle{Real-time procedural generation ofpseudo
  infinite'cities}. In \bibinfo{booktitle}{\emph{Proceedings of the 1st
  international conference on Computer graphics and interactive techniques in
  Australasia and South East Asia}}. \bibinfo{pages}{87--ff}.
\newblock


\bibitem[Grover et~al\mbox{.}(2019)]%
        {Grover2019GraphiteIG}
\bibfield{author}{\bibinfo{person}{Aditya Grover}, \bibinfo{person}{Aaron
  Zweig}, {and} \bibinfo{person}{S. Ermon}.} \bibinfo{year}{2019}\natexlab{}.
\newblock \showarticletitle{Graphite: Iterative Generative Modeling of Graphs}.
\newblock \bibinfo{journal}{\emph{ArXiv}}  \bibinfo{volume}{abs/1803.10459}
  (\bibinfo{year}{2019}).
\newblock


\bibitem[Guimaraes et~al\mbox{.}(2017)]%
        {Guimaraes2017ObjectiveReinforcedGA}
\bibfield{author}{\bibinfo{person}{G.~L. Guimaraes},
  \bibinfo{person}{Benjam{\'i}n S{\'a}nchez-Lengeling}, \bibinfo{person}{Pedro
  Luis~Cunha Farias}, {and} \bibinfo{person}{Al{\'a}n Aspuru-Guzik}.}
  \bibinfo{year}{2017}\natexlab{}.
\newblock \showarticletitle{Objective-Reinforced Generative Adversarial
  Networks (ORGAN) for Sequence Generation Models}.
\newblock \bibinfo{journal}{\emph{ArXiv}}  \bibinfo{volume}{abs/1705.10843}
  (\bibinfo{year}{2017}).
\newblock


\bibitem[Hartmann et~al\mbox{.}(2017)]%
        {hartmann2017streetgan}
\bibfield{author}{\bibinfo{person}{Stefan Hartmann}, \bibinfo{person}{Michael
  Weinmann}, \bibinfo{person}{Raoul Wessel}, {and} \bibinfo{person}{Reinhard
  Klein}.} \bibinfo{year}{2017}\natexlab{}.
\newblock \showarticletitle{Streetgan: Towards road network synthesis with
  generative adversarial networks}.
\newblock  (\bibinfo{year}{2017}).
\newblock


\bibitem[He et~al\mbox{.}(2016)]%
        {he2016deep}
\bibfield{author}{\bibinfo{person}{Kaiming He}, \bibinfo{person}{Xiangyu
  Zhang}, \bibinfo{person}{Shaoqing Ren}, {and} \bibinfo{person}{Jian Sun}.}
  \bibinfo{year}{2016}\natexlab{}.
\newblock \showarticletitle{Deep residual learning for image recognition}. In
  \bibinfo{booktitle}{\emph{Proceedings of the IEEE conference on computer
  vision and pattern recognition}}. \bibinfo{pages}{770--778}.
\newblock


\bibitem[Ho et~al\mbox{.}(2020)]%
        {ho2020axial}
\bibfield{author}{\bibinfo{person}{Jonathan Ho}, \bibinfo{person}{Nal
  Kalchbrenner}, \bibinfo{person}{Dirk Weissenborn}, {and} \bibinfo{person}{Tim
  Salimans}.} \bibinfo{year}{2020}\natexlab{}.
\newblock \bibinfo{title}{Axial Attention in Multidimensional Transformers}.
\newblock
\newblock
\urldef\tempurl%
\url{https://openreview.net/forum?id=H1e5GJBtDr}
\showURL{%
\tempurl}


\bibitem[Honda et~al\mbox{.}(2020)]%
        {honda2020graph}
\bibfield{author}{\bibinfo{person}{Shion Honda}, \bibinfo{person}{Hirotaka
  Akita}, \bibinfo{person}{Katsuhiko Ishiguro}, \bibinfo{person}{Toshiki
  Nakanishi}, {and} \bibinfo{person}{Kenta Oono}.}
  \bibinfo{year}{2020}\natexlab{}.
\newblock \bibinfo{title}{Graph Residual Flow for Molecular Graph Generation}.
\newblock
\newblock
\urldef\tempurl%
\url{https://openreview.net/forum?id=SyepHTNFDS}
\showURL{%
\tempurl}


\bibitem[Jain et~al\mbox{.}(2020)]%
        {jain2020lmconv}
\bibfield{author}{\bibinfo{person}{Ajay Jain}, \bibinfo{person}{Pieter Abbeel},
  {and} \bibinfo{person}{Deepak Pathak}.} \bibinfo{year}{2020}\natexlab{}.
\newblock \showarticletitle{Locally Masked Convolution for Autoregressive
  Models}. In \bibinfo{booktitle}{\emph{Conference on Uncertainty in Artificial
  Intelligence (UAI)}}.
\newblock


\bibitem[Jiang(2009)]%
        {jiang2009ranking}
\bibfield{author}{\bibinfo{person}{Bin Jiang}.}
  \bibinfo{year}{2009}\natexlab{}.
\newblock \showarticletitle{Ranking spaces for predicting human movement in an
  urban environment}.
\newblock \bibinfo{journal}{\emph{International Journal of Geographical
  Information Science}} \bibinfo{volume}{23}, \bibinfo{number}{7}
  (\bibinfo{year}{2009}), \bibinfo{pages}{823--837}.
\newblock


\bibitem[Karras et~al\mbox{.}(2020)]%
        {Karras2020ada}
\bibfield{author}{\bibinfo{person}{Tero Karras}, \bibinfo{person}{Miika
  Aittala}, \bibinfo{person}{Janne Hellsten}, \bibinfo{person}{Samuli Laine},
  \bibinfo{person}{Jaakko Lehtinen}, {and} \bibinfo{person}{Timo Aila}.}
  \bibinfo{year}{2020}\natexlab{}.
\newblock \showarticletitle{Training Generative Adversarial Networks with
  Limited Data}. In \bibinfo{booktitle}{\emph{Proc. NeurIPS}}.
\newblock


\bibitem[Karras et~al\mbox{.}(2021)]%
        {karras2021aliasfree}
\bibfield{author}{\bibinfo{person}{Tero Karras}, \bibinfo{person}{Miika
  Aittala}, \bibinfo{person}{Samuli Laine}, \bibinfo{person}{Erik Härkönen},
  \bibinfo{person}{Janne Hellsten}, \bibinfo{person}{Jaakko Lehtinen}, {and}
  \bibinfo{person}{Timo Aila}.} \bibinfo{year}{2021}\natexlab{}.
\newblock \showarticletitle{Alias-Free Generative Adversarial Networks}. In
  \bibinfo{booktitle}{\emph{NeurIPS}}.
\newblock


\bibitem[Katharopoulos et~al\mbox{.}(2020)]%
        {katharopoulos_et_al_2020}
\bibfield{author}{\bibinfo{person}{A. Katharopoulos}, \bibinfo{person}{A.
  Vyas}, \bibinfo{person}{N. Pappas}, {and} \bibinfo{person}{F. Fleuret}.}
  \bibinfo{year}{2020}\natexlab{}.
\newblock \showarticletitle{Transformers are RNNs: Fast Autoregressive
  Transformers with Linear Attention}. In \bibinfo{booktitle}{\emph{Proceedings
  of the International Conference on Machine Learning (ICML)}}.
\newblock


\bibitem[Kelly and McCabe(2007)]%
        {kelly2007citygen}
\bibfield{author}{\bibinfo{person}{George Kelly} {and} \bibinfo{person}{Hugh
  McCabe}.} \bibinfo{year}{2007}\natexlab{}.
\newblock \showarticletitle{Citygen: An interactive system for procedural city
  generation}. In \bibinfo{booktitle}{\emph{Fifth International Conference on
  Game Design and Technology}}. \bibinfo{pages}{8--16}.
\newblock


\bibitem[Kim et~al\mbox{.}(2018)]%
        {Kim2018}
\bibfield{author}{\bibinfo{person}{Joon-Seok Kim}, \bibinfo{person}{Hamdi
  Kavak}, {and} \bibinfo{person}{Andrew Crooks}.}
  \bibinfo{year}{2018}\natexlab{}.
\newblock \showarticletitle{Procedural City Generation beyond Game
  Development}.
\newblock  \bibinfo{volume}{10}, \bibinfo{number}{2} (\bibinfo{date}{nov}
  \bibinfo{year}{2018}), \bibinfo{pages}{34–41}.
\newblock
\urldef\tempurl%
\url{https://doi.org/10.1145/3292390.3292397}
\showDOI{\tempurl}


\bibitem[Kingma et~al\mbox{.}(2016)]%
        {KingmaIAF}
\bibfield{author}{\bibinfo{person}{Durk~P Kingma}, \bibinfo{person}{Tim
  Salimans}, \bibinfo{person}{Rafal Jozefowicz}, \bibinfo{person}{Xi Chen},
  \bibinfo{person}{Ilya Sutskever}, {and} \bibinfo{person}{Max Welling}.}
  \bibinfo{year}{2016}\natexlab{}.
\newblock \showarticletitle{Improved Variational Inference with Inverse
  Autoregressive Flow}. In \bibinfo{booktitle}{\emph{Advances in Neural
  Information Processing Systems}}, \bibfield{editor}{\bibinfo{person}{D.~Lee},
  \bibinfo{person}{M.~Sugiyama}, \bibinfo{person}{U.~Luxburg},
  \bibinfo{person}{I.~Guyon}, {and} \bibinfo{person}{R.~Garnett}} (Eds.),
  Vol.~\bibinfo{volume}{29}. \bibinfo{publisher}{Curran Associates, Inc.}
\newblock
\urldef\tempurl%
\url{https://proceedings.neurips.cc/paper/2016/file/ddeebdeefdb7e7e7a697e1c3e3d8ef54-Paper.pdf}
\showURL{%
\tempurl}


\bibitem[Kingma and Welling(2014)]%
        {KingmaWellingVAE}
\bibfield{author}{\bibinfo{person}{Diederik~P. Kingma} {and}
  \bibinfo{person}{Max Welling}.} \bibinfo{year}{2014}\natexlab{}.
\newblock \showarticletitle{Auto-Encoding Variational Bayes}. In
  \bibinfo{booktitle}{\emph{2nd International Conference on Learning
  Representations, {ICLR} 2014, Banff, AB, Canada, April 14-16, 2014,
  Conference Track Proceedings}}, \bibfield{editor}{\bibinfo{person}{Yoshua
  Bengio} {and} \bibinfo{person}{Yann LeCun}} (Eds.).
\newblock
\urldef\tempurl%
\url{http://arxiv.org/abs/1312.6114}
\showURL{%
\tempurl}


\bibitem[Krecklau et~al\mbox{.}(2012)]%
        {krecklau2012procedural}
\bibfield{author}{\bibinfo{person}{Lars Krecklau}, \bibinfo{person}{Christopher
  Manthei}, {and} \bibinfo{person}{Leif Kobbelt}.}
  \bibinfo{year}{2012}\natexlab{}.
\newblock \showarticletitle{Procedural interpolation of historical city maps}.
  In \bibinfo{booktitle}{\emph{Computer Graphics Forum}},
  Vol.~\bibinfo{volume}{31}. Wiley Online Library, \bibinfo{pages}{691--700}.
\newblock


\bibitem[Kusner et~al\mbox{.}(2017)]%
        {kusner2016GVAE}
\bibfield{author}{\bibinfo{person}{Matt~J. Kusner}, \bibinfo{person}{Brooks
  Paige}, {and} \bibinfo{person}{Jos\'{e}~Miguel Hern\'{a}ndez-Lobato}.}
  \bibinfo{year}{2017}\natexlab{}.
\newblock \showarticletitle{Grammar Variational Autoencoder}. In
  \bibinfo{booktitle}{\emph{Proceedings of the 34th International Conference on
  Machine Learning - Volume 70}} (Sydney, NSW, Australia)
  \emph{(\bibinfo{series}{ICML'17})}. \bibinfo{publisher}{JMLR.org},
  \bibinfo{pages}{1945–1954}.
\newblock


\bibitem[Lechner et~al\mbox{.}(2006)]%
        {lechner2006procedural}
\bibfield{author}{\bibinfo{person}{Thomas Lechner}, \bibinfo{person}{Pin Ren},
  \bibinfo{person}{Ben Watson}, \bibinfo{person}{Craig Brozefski}, {and}
  \bibinfo{person}{Uri Wilenski}.} \bibinfo{year}{2006}\natexlab{}.
\newblock \showarticletitle{Procedural modeling of urban land use}.
\newblock In \bibinfo{booktitle}{\emph{ACM SIGGRAPH 2006 Research posters}}.
  \bibinfo{pages}{135--es}.
\newblock


\bibitem[{Lecun} et~al\mbox{.}(1998)]%
        {LecunnGradient}
\bibfield{author}{\bibinfo{person}{Y. {Lecun}}, \bibinfo{person}{L. {Bottou}},
  \bibinfo{person}{Y. {Bengio}}, {and} \bibinfo{person}{P. {Haffner}}.}
  \bibinfo{year}{1998}\natexlab{}.
\newblock \showarticletitle{Gradient-based learning applied to document
  recognition}.
\newblock \bibinfo{journal}{\emph{Proc. IEEE}} \bibinfo{volume}{86},
  \bibinfo{number}{11} (\bibinfo{year}{1998}), \bibinfo{pages}{2278--2324}.
\newblock
\urldef\tempurl%
\url{https://doi.org/10.1109/5.726791}
\showDOI{\tempurl}


\bibitem[Li et~al\mbox{.}(2021)]%
        {li2021graph}
\bibfield{author}{\bibinfo{person}{Jia Li}, \bibinfo{person}{Jianwei Yu},
  \bibinfo{person}{Da-Cheng Juan}, \bibinfo{person}{HAN Zhichao},
  \bibinfo{person}{Arjun Gopalan}, \bibinfo{person}{Hong Cheng}, {and}
  \bibinfo{person}{Andrew Tomkins}.} \bibinfo{year}{2021}\natexlab{}.
\newblock \bibinfo{title}{Graph Autoencoders with Deconvolutional Networks}.
\newblock
\newblock
\urldef\tempurl%
\url{https://openreview.net/forum?id=ohz3OEhVcs}
\showURL{%
\tempurl}


\bibitem[Li et~al\mbox{.}(2018)]%
        {li2018learning}
\bibfield{author}{\bibinfo{person}{Yujia Li}, \bibinfo{person}{Oriol Vinyals},
  \bibinfo{person}{Chris Dyer}, \bibinfo{person}{Razvan Pascanu}, {and}
  \bibinfo{person}{Peter Battaglia}.} \bibinfo{year}{2018}\natexlab{}.
\newblock \bibinfo{title}{Learning Deep Generative Models of Graphs}.
\newblock
\newblock
\urldef\tempurl%
\url{https://openreview.net/forum?id=Hy1d-ebAb}
\showURL{%
\tempurl}


\bibitem[Li et~al\mbox{.}(2016)]%
        {li2016gated}
\bibfield{author}{\bibinfo{person}{Yujia Li}, \bibinfo{person}{Richard Zemel},
  \bibinfo{person}{Marc Brockschmidt}, {and} \bibinfo{person}{Daniel Tarlow}.}
  \bibinfo{year}{2016}\natexlab{}.
\newblock \showarticletitle{Gated Graph Sequence Neural Networks}. In
  \bibinfo{booktitle}{\emph{Proceedings of ICLR'16}
  (\bibinfo{edition}{proceedings of iclr'16} ed.)}.
\newblock
\urldef\tempurl%
\url{https://www.microsoft.com/en-us/research/publication/gated-graph-sequence-neural-networks/}
\showURL{%
\tempurl}


\bibitem[Liao et~al\mbox{.}(2019)]%
        {liao2019gran}
\bibfield{author}{\bibinfo{person}{Renjie Liao}, \bibinfo{person}{Yujia Li},
  \bibinfo{person}{Yang Song}, \bibinfo{person}{Shenlong Wang},
  \bibinfo{person}{Will Hamilton}, \bibinfo{person}{David~K Duvenaud},
  \bibinfo{person}{Raquel Urtasun}, {and} \bibinfo{person}{Richard Zemel}.}
  \bibinfo{year}{2019}\natexlab{}.
\newblock \showarticletitle{Efficient Graph Generation with Graph Recurrent
  Attention Networks}. In \bibinfo{booktitle}{\emph{Advances in Neural
  Information Processing Systems}},
  \bibfield{editor}{\bibinfo{person}{H.~Wallach},
  \bibinfo{person}{H.~Larochelle}, \bibinfo{person}{A.~Beygelzimer},
  \bibinfo{person}{F.~d\textquotesingle Alch\'{e}-Buc},
  \bibinfo{person}{E.~Fox}, {and} \bibinfo{person}{R.~Garnett}} (Eds.),
  Vol.~\bibinfo{volume}{32}. \bibinfo{publisher}{Curran Associates, Inc.}
\newblock
\urldef\tempurl%
\url{https://proceedings.neurips.cc/paper/2019/file/d0921d442ee91b896ad95059d13df618-Paper.pdf}
\showURL{%
\tempurl}


\bibitem[Lin et~al\mbox{.}(2017)]%
        {lin2017structured}
\bibfield{author}{\bibinfo{person}{Zhouhan Lin}, \bibinfo{person}{Minwei Feng},
  \bibinfo{person}{Cicero Nogueira~dos Santos}, \bibinfo{person}{Mo Yu},
  \bibinfo{person}{Bing Xiang}, \bibinfo{person}{Bowen Zhou}, {and}
  \bibinfo{person}{Yoshua Bengio}.} \bibinfo{year}{2017}\natexlab{}.
\newblock \showarticletitle{A structured self-attentive sentence embedding}.
\newblock \bibinfo{journal}{\emph{arXiv preprint arXiv:1703.03130}}
  (\bibinfo{year}{2017}).
\newblock


\bibitem[Maal{\o}e et~al\mbox{.}(2019)]%
        {Maale2019BIVAAV}
\bibfield{author}{\bibinfo{person}{Lars Maal{\o}e}, \bibinfo{person}{M.
  Fraccaro}, \bibinfo{person}{Valentin Li{\'e}vin}, {and} \bibinfo{person}{O.
  Winther}.} \bibinfo{year}{2019}\natexlab{}.
\newblock \showarticletitle{BIVA: A Very Deep Hierarchy of Latent Variables for
  Generative Modeling}. In \bibinfo{booktitle}{\emph{NeurIPS}}.
\newblock


\bibitem[Mas et~al\mbox{.}(2020)]%
        {mas2020simulating}
\bibfield{author}{\bibinfo{person}{Albert Mas}, \bibinfo{person}{Ignacio
  Martin}, {and} \bibinfo{person}{Gustavo Patow}.}
  \bibinfo{year}{2020}\natexlab{}.
\newblock \showarticletitle{Simulating the Evolution of Ancient Fortified
  Cities}. In \bibinfo{booktitle}{\emph{Computer Graphics Forum}},
  Vol.~\bibinfo{volume}{39}. Wiley Online Library, \bibinfo{pages}{650--671}.
\newblock


\bibitem[Mi et~al\mbox{.}(2021)]%
        {mi2021hdmapgen}
\bibfield{author}{\bibinfo{person}{Lu Mi}, \bibinfo{person}{Hang Zhao},
  \bibinfo{person}{Charlie Nash}, \bibinfo{person}{Xiaohan Jin},
  \bibinfo{person}{Jiyang Gao}, \bibinfo{person}{Chen Sun},
  \bibinfo{person}{Cordelia Schmid}, \bibinfo{person}{Nir Shavit},
  \bibinfo{person}{Yuning Chai}, {and} \bibinfo{person}{Dragomir Anguelov}.}
  \bibinfo{year}{2021}\natexlab{}.
\newblock \showarticletitle{Hdmapgen: A hierarchical graph generative model of
  high definition maps}. In \bibinfo{booktitle}{\emph{Proc. CVPR}}.
  \bibinfo{pages}{4227--4236}.
\newblock


\bibitem[Nash et~al\mbox{.}(2020)]%
        {pmlr-v119-nash20a}
\bibfield{author}{\bibinfo{person}{Charlie Nash}, \bibinfo{person}{Yaroslav
  Ganin}, \bibinfo{person}{S.~M.~Ali Eslami}, {and} \bibinfo{person}{Peter
  Battaglia}.} \bibinfo{year}{2020}\natexlab{}.
\newblock \showarticletitle{{P}oly{G}en: An Autoregressive Generative Model of
  3{D} Meshes}. In \bibinfo{booktitle}{\emph{Proceedings of the 37th
  International Conference on Machine Learning}}
  \emph{(\bibinfo{series}{Proceedings of Machine Learning Research},
  Vol.~\bibinfo{volume}{119})}, \bibfield{editor}{\bibinfo{person}{Hal~Daumé
  III} {and} \bibinfo{person}{Aarti Singh}} (Eds.). \bibinfo{publisher}{PMLR},
  \bibinfo{pages}{7220--7229}.
\newblock
\urldef\tempurl%
\url{http://proceedings.mlr.press/v119/nash20a.html}
\showURL{%
\tempurl}


\bibitem[Nishida et~al\mbox{.}(2016)]%
        {nishida2016example}
\bibfield{author}{\bibinfo{person}{Gen Nishida}, \bibinfo{person}{Ignacio
  Garcia-Dorado}, {and} \bibinfo{person}{Daniel~G Aliaga}.}
  \bibinfo{year}{2016}\natexlab{}.
\newblock \showarticletitle{Example-driven procedural urban roads}. In
  \bibinfo{booktitle}{\emph{Computer Graphics Forum}},
  Vol.~\bibinfo{volume}{35}. Wiley Online Library, \bibinfo{pages}{5--17}.
\newblock


\bibitem[Oord et~al\mbox{.}(2016a)]%
        {OordRecurrent}
\bibfield{author}{\bibinfo{person}{Aaron~Van Oord}, \bibinfo{person}{Nal
  Kalchbrenner}, {and} \bibinfo{person}{Koray Kavukcuoglu}.}
  \bibinfo{year}{2016}\natexlab{a}.
\newblock \showarticletitle{Pixel Recurrent Neural Networks}. In
  \bibinfo{booktitle}{\emph{Proceedings of The 33rd International Conference on
  Machine Learning}} \emph{(\bibinfo{series}{Proceedings of Machine Learning
  Research}, Vol.~\bibinfo{volume}{48})},
  \bibfield{editor}{\bibinfo{person}{Maria~Florina Balcan} {and}
  \bibinfo{person}{Kilian~Q. Weinberger}} (Eds.). \bibinfo{publisher}{PMLR},
  \bibinfo{address}{New York, New York, USA}, \bibinfo{pages}{1747--1756}.
\newblock
\urldef\tempurl%
\url{http://proceedings.mlr.press/v48/oord16.html}
\showURL{%
\tempurl}


\bibitem[Oord et~al\mbox{.}(2016b)]%
        {OordPixelCNN}
\bibfield{author}{\bibinfo{person}{A\"{a}ron van~den Oord},
  \bibinfo{person}{Nal Kalchbrenner}, \bibinfo{person}{Oriol Vinyals},
  \bibinfo{person}{Lasse Espeholt}, \bibinfo{person}{Alex Graves}, {and}
  \bibinfo{person}{Koray Kavukcuoglu}.} \bibinfo{year}{2016}\natexlab{b}.
\newblock \showarticletitle{Conditional Image Generation with PixelCNN
  Decoders}. In \bibinfo{booktitle}{\emph{Proceedings of the 30th International
  Conference on Neural Information Processing Systems}} (Barcelona, Spain)
  \emph{(\bibinfo{series}{NIPS'16})}. \bibinfo{publisher}{Curran Associates
  Inc.}, \bibinfo{address}{Red Hook, NY, USA}, \bibinfo{pages}{4797–4805}.
\newblock
\showISBNx{9781510838819}


\bibitem[Owaki and Machida(2020)]%
        {owaki2020roadnetgan}
\bibfield{author}{\bibinfo{person}{Takashi Owaki} {and}
  \bibinfo{person}{Takashi Machida}.} \bibinfo{year}{2020}\natexlab{}.
\newblock \showarticletitle{RoadNetGAN: Generating Road Networks in Planar
  Graph Representation}. In \bibinfo{booktitle}{\emph{International Conference
  on Neural Information Processing}}. Springer, \bibinfo{pages}{535--543}.
\newblock


\bibitem[Page et~al\mbox{.}(1999)]%
        {page1999pagerank}
\bibfield{author}{\bibinfo{person}{Lawrence Page}, \bibinfo{person}{Sergey
  Brin}, \bibinfo{person}{Rajeev Motwani}, {and} \bibinfo{person}{Terry
  Winograd}.} \bibinfo{year}{1999}\natexlab{}.
\newblock \bibinfo{booktitle}{\emph{The PageRank citation ranking: Bringing
  order to the web.}}
\newblock \bibinfo{type}{{T}echnical {R}eport}. \bibinfo{institution}{Stanford
  InfoLab}.
\newblock


\bibitem[Para et~al\mbox{.}(2020)]%
        {para2020generative}
\bibfield{author}{\bibinfo{person}{Wamiq Para}, \bibinfo{person}{Paul
  Guerrero}, \bibinfo{person}{Tom Kelly}, \bibinfo{person}{Leonidas Guibas},
  {and} \bibinfo{person}{Peter Wonka}.} \bibinfo{year}{2020}\natexlab{}.
\newblock \showarticletitle{Generative Layout Modeling using Constraint
  Graphs}.
\newblock \bibinfo{journal}{\emph{arXiv preprint arXiv:2011.13417}}
  (\bibinfo{year}{2020}).
\newblock


\bibitem[Parikh et~al\mbox{.}(2016)]%
        {parikh-etal-2016-decomposable}
\bibfield{author}{\bibinfo{person}{Ankur Parikh}, \bibinfo{person}{Oscar
  T{\"a}ckstr{\"o}m}, \bibinfo{person}{Dipanjan Das}, {and}
  \bibinfo{person}{Jakob Uszkoreit}.} \bibinfo{year}{2016}\natexlab{}.
\newblock \showarticletitle{A Decomposable Attention Model for Natural Language
  Inference}. In \bibinfo{booktitle}{\emph{Proceedings of the 2016 Conference
  on Empirical Methods in Natural Language Processing}}.
  \bibinfo{publisher}{Association for Computational Linguistics},
  \bibinfo{address}{Austin, Texas}, \bibinfo{pages}{2249--2255}.
\newblock
\urldef\tempurl%
\url{https://doi.org/10.18653/v1/D16-1244}
\showDOI{\tempurl}


\bibitem[Parish and M{\"u}ller(2001)]%
        {Parish:2001:PMC}
\bibfield{author}{\bibinfo{person}{Yoav I.~H. Parish} {and}
  \bibinfo{person}{Pascal M{\"u}ller}.} \bibinfo{year}{2001}\natexlab{}.
\newblock \showarticletitle{Procedural Modeling of Cities}. In
  \bibinfo{booktitle}{\emph{Proceedings of SIGGRAPH 2001}}.
  \bibinfo{pages}{301--308}.
\newblock
\urldef\tempurl%
\url{https://doi.org/10.1145/383259.383292}
\showDOI{\tempurl}


\bibitem[Parmar et~al\mbox{.}(2018)]%
        {parmar2018image}
\bibfield{author}{\bibinfo{person}{Niki Parmar}, \bibinfo{person}{Ashish
  Vaswani}, \bibinfo{person}{Jakob Uszkoreit}, \bibinfo{person}{Lukasz Kaiser},
  \bibinfo{person}{Noam Shazeer}, \bibinfo{person}{Alexander Ku}, {and}
  \bibinfo{person}{Dustin Tran}.} \bibinfo{year}{2018}\natexlab{}.
\newblock \showarticletitle{Image transformer}. In
  \bibinfo{booktitle}{\emph{International Conference on Machine Learning}}.
  PMLR, \bibinfo{pages}{4055--4064}.
\newblock


\bibitem[Peng et~al\mbox{.}(2014)]%
        {peng2014computing}
\bibfield{author}{\bibinfo{person}{Chi-Han Peng}, \bibinfo{person}{Yong-Liang
  Yang}, {and} \bibinfo{person}{Peter Wonka}.} \bibinfo{year}{2014}\natexlab{}.
\newblock \showarticletitle{Computing layouts with deformable templates}.
\newblock \bibinfo{journal}{\emph{ACM Transactions on Graphics (TOG)}}
  \bibinfo{volume}{33}, \bibinfo{number}{4} (\bibinfo{year}{2014}),
  \bibinfo{pages}{1--11}.
\newblock


\bibitem[Prusinkiewicz(1986)]%
        {prusinkiewicz1986graphical}
\bibfield{author}{\bibinfo{person}{Przemyslaw Prusinkiewicz}.}
  \bibinfo{year}{1986}\natexlab{}.
\newblock \showarticletitle{Graphical applications of L-systems}. In
  \bibinfo{booktitle}{\emph{Proceedings of graphics interface}},
  Vol.~\bibinfo{volume}{86}. \bibinfo{pages}{247--253}.
\newblock


\bibitem[Pueyo et~al\mbox{.}(2020)]%
        {pueyo2020shrinking}
\bibfield{author}{\bibinfo{person}{Oriol Pueyo}, \bibinfo{person}{Albert
  Sabri{\`a}}, \bibinfo{person}{Xavier Pueyo}, \bibinfo{person}{Gustavo Patow},
  {and} \bibinfo{person}{Michael Wimmer}.} \bibinfo{year}{2020}\natexlab{}.
\newblock \showarticletitle{Shrinking city layouts}.
\newblock \bibinfo{journal}{\emph{Computers \& Graphics}}  \bibinfo{volume}{86}
  (\bibinfo{year}{2020}), \bibinfo{pages}{15--26}.
\newblock


\bibitem[Ramesh et~al\mbox{.}(2022)]%
        {DallE2}
\bibfield{author}{\bibinfo{person}{Aditya Ramesh}, \bibinfo{person}{Prafulla
  Dhariwal}, \bibinfo{person}{Alex Nichol}, \bibinfo{person}{Casey Chu}, {and}
  \bibinfo{person}{Mark Chen}.} \bibinfo{year}{2022}\natexlab{}.
\newblock \bibinfo{title}{Hierarchical Text-Conditional Image Generation with
  CLIP Latents}.
\newblock
\newblock


\bibitem[Ramesh et~al\mbox{.}(2021)]%
        {ramesh2021zeroshot}
\bibfield{author}{\bibinfo{person}{Aditya Ramesh}, \bibinfo{person}{Mikhail
  Pavlov}, \bibinfo{person}{Gabriel Goh}, \bibinfo{person}{Scott Gray},
  \bibinfo{person}{Chelsea Voss}, \bibinfo{person}{Alec Radford},
  \bibinfo{person}{Mark Chen}, {and} \bibinfo{person}{Ilya Sutskever}.}
  \bibinfo{year}{2021}\natexlab{}.
\newblock \bibinfo{title}{Zero-Shot Text-to-Image Generation}.
\newblock
\newblock
\showeprint[arxiv]{2102.12092}~[cs.CV]


\bibitem[Ranganath et~al\mbox{.}(2016)]%
        {Ranganath2016HierarchicalVM}
\bibfield{author}{\bibinfo{person}{R. Ranganath}, \bibinfo{person}{Dustin
  Tran}, {and} \bibinfo{person}{D. Blei}.} \bibinfo{year}{2016}\natexlab{}.
\newblock \showarticletitle{Hierarchical Variational Models}.
\newblock \bibinfo{journal}{\emph{ArXiv}}  \bibinfo{volume}{abs/1511.02386}
  (\bibinfo{year}{2016}).
\newblock


\bibitem[Razavi et~al\mbox{.}(2019)]%
        {OordVAE2}
\bibfield{author}{\bibinfo{person}{Ali Razavi}, \bibinfo{person}{Aaron van~den
  Oord}, {and} \bibinfo{person}{Oriol Vinyals}.}
  \bibinfo{year}{2019}\natexlab{}.
\newblock \showarticletitle{Generating Diverse High-Fidelity Images with
  VQ-VAE-2}. In \bibinfo{booktitle}{\emph{Advances in Neural Information
  Processing Systems}}, \bibfield{editor}{\bibinfo{person}{H.~Wallach},
  \bibinfo{person}{H.~Larochelle}, \bibinfo{person}{A.~Beygelzimer},
  \bibinfo{person}{F.~d\textquotesingle Alch\'{e}-Buc},
  \bibinfo{person}{E.~Fox}, {and} \bibinfo{person}{R.~Garnett}} (Eds.),
  Vol.~\bibinfo{volume}{32}. \bibinfo{publisher}{Curran Associates, Inc.}
\newblock
\urldef\tempurl%
\url{https://proceedings.neurips.cc/paper/2019/file/5f8e2fa1718d1bbcadf1cd9c7a54fb8c-Paper.pdf}
\showURL{%
\tempurl}


\bibitem[Ronneberger et~al\mbox{.}(2015)]%
        {OlafUNET}
\bibfield{author}{\bibinfo{person}{Olaf Ronneberger}, \bibinfo{person}{Philipp
  Fischer}, {and} \bibinfo{person}{Thomas Brox}.}
  \bibinfo{year}{2015}\natexlab{}.
\newblock \showarticletitle{U-Net: Convolutional Networks for Biomedical Image
  Segmentation}. In \bibinfo{booktitle}{\emph{Medical Image Computing and
  Computer-Assisted Intervention -- MICCAI 2015}},
  \bibfield{editor}{\bibinfo{person}{Nassir Navab}, \bibinfo{person}{Joachim
  Hornegger}, \bibinfo{person}{William~M. Wells}, {and}
  \bibinfo{person}{Alejandro~F. Frangi}} (Eds.). \bibinfo{publisher}{Springer
  International Publishing}, \bibinfo{address}{Cham},
  \bibinfo{pages}{234--241}.
\newblock


\bibitem[Salimans et~al\mbox{.}(2017)]%
        {Salimans2017PixeCNN}
\bibfield{author}{\bibinfo{person}{Tim Salimans}, \bibinfo{person}{Andrej
  Karpathy}, \bibinfo{person}{Xi Chen}, {and} \bibinfo{person}{Diederik~P.
  Kingma}.} \bibinfo{year}{2017}\natexlab{}.
\newblock \showarticletitle{PixelCNN++: A PixelCNN Implementation with
  Discretized Logistic Mixture Likelihood and Other Modifications}. In
  \bibinfo{booktitle}{\emph{ICLR}}.
\newblock


\bibitem[Shi* et~al\mbox{.}(2020)]%
        {Shi*2020GraphAF}
\bibfield{author}{\bibinfo{person}{Chence Shi*}, \bibinfo{person}{Minkai Xu*},
  \bibinfo{person}{Zhaocheng Zhu}, \bibinfo{person}{Weinan Zhang},
  \bibinfo{person}{Ming Zhang}, {and} \bibinfo{person}{Jian Tang}.}
  \bibinfo{year}{2020}\natexlab{}.
\newblock \showarticletitle{GraphAF: a Flow-based Autoregressive Model for
  Molecular Graph Generation}. In \bibinfo{booktitle}{\emph{International
  Conference on Learning Representations}}.
\newblock
\urldef\tempurl%
\url{https://openreview.net/forum?id=S1esMkHYPr}
\showURL{%
\tempurl}


\bibitem[Simonovsky and Komodakis(2018)]%
        {simonovsky2018graphvae}
\bibfield{author}{\bibinfo{person}{Martin Simonovsky} {and}
  \bibinfo{person}{Nikos Komodakis}.} \bibinfo{year}{2018}\natexlab{}.
\newblock \bibinfo{title}{Graph{VAE}: Towards Generation of Small Graphs Using
  Variational Autoencoders}.
\newblock
\newblock
\urldef\tempurl%
\url{https://openreview.net/forum?id=SJlhPMWAW}
\showURL{%
\tempurl}


\bibitem[Smelik et~al\mbox{.}(2014)]%
        {smelik2014survey}
\bibfield{author}{\bibinfo{person}{Ruben~M Smelik}, \bibinfo{person}{Tim
  Tutenel}, \bibinfo{person}{Rafael Bidarra}, {and} \bibinfo{person}{Bedrich
  Benes}.} \bibinfo{year}{2014}\natexlab{}.
\newblock \showarticletitle{A survey on procedural modelling for virtual
  worlds}. In \bibinfo{booktitle}{\emph{Computer Graphics Forum}},
  Vol.~\bibinfo{volume}{33}. Wiley Online Library, \bibinfo{pages}{31--50}.
\newblock


\bibitem[S{\o}nderby et~al\mbox{.}(2016)]%
        {Snderby2016LadderVA}
\bibfield{author}{\bibinfo{person}{C. S{\o}nderby}, \bibinfo{person}{T. Raiko},
  \bibinfo{person}{Lars Maal{\o}e}, \bibinfo{person}{S{\o}ren~Kaae
  S{\o}nderby}, {and} \bibinfo{person}{O. Winther}.}
  \bibinfo{year}{2016}\natexlab{}.
\newblock \showarticletitle{Ladder Variational Autoencoders}. In
  \bibinfo{booktitle}{\emph{NIPS}}.
\newblock


\bibitem[Song and Whitehead(2019)]%
        {song2019townsim}
\bibfield{author}{\bibinfo{person}{Asiiah Song} {and} \bibinfo{person}{Jim
  Whitehead}.} \bibinfo{year}{2019}\natexlab{}.
\newblock \showarticletitle{TownSim: Agent-based city evolution for
  naturalistic road network generation}. In
  \bibinfo{booktitle}{\emph{Proceedings of the 14th International Conference on
  the Foundations of Digital Games}}. \bibinfo{pages}{1--9}.
\newblock


\bibitem[Spencer(2009)]%
        {Spencer2019}
\bibfield{author}{\bibinfo{person}{Douglas Spencer}.}
  \bibinfo{year}{2009}\natexlab{}.
\newblock \showarticletitle{Cities and Complexity: Understanding Cities with
  Cellular Automata, Agent-Based Models, and Fractals}.
\newblock \bibinfo{journal}{\emph{The Journal of Architecture}}
  \bibinfo{volume}{14}, \bibinfo{number}{3} (\bibinfo{year}{2009}),
  \bibinfo{pages}{446--450}.
\newblock
\urldef\tempurl%
\url{https://doi.org/10.1080/13602360903028044}
\showDOI{\tempurl}


\bibitem[Sun et~al\mbox{.}(2002)]%
        {sun2002template}
\bibfield{author}{\bibinfo{person}{Jing Sun}, \bibinfo{person}{Xiaobo Yu},
  \bibinfo{person}{George Baciu}, {and} \bibinfo{person}{Mark Green}.}
  \bibinfo{year}{2002}\natexlab{}.
\newblock \showarticletitle{Template-based generation of road networks for
  virtual city modeling}. In \bibinfo{booktitle}{\emph{Proceedings of the ACM
  symposium on Virtual reality software and technology}}.
  \bibinfo{pages}{33--40}.
\newblock


\bibitem[Tomczak and Welling(2018)]%
        {TomczakVampPrior}
\bibfield{author}{\bibinfo{person}{Jakub Tomczak} {and} \bibinfo{person}{Max
  Welling}.} \bibinfo{year}{2018}\natexlab{}.
\newblock \showarticletitle{VAE with a VampPrior}. In
  \bibinfo{booktitle}{\emph{Proceedings of the Twenty-First International
  Conference on Artificial Intelligence and Statistics}}
  \emph{(\bibinfo{series}{Proceedings of Machine Learning Research},
  Vol.~\bibinfo{volume}{84})}, \bibfield{editor}{\bibinfo{person}{Amos Storkey}
  {and} \bibinfo{person}{Fernando Perez-Cruz}} (Eds.).
  \bibinfo{publisher}{PMLR}, \bibinfo{pages}{1214--1223}.
\newblock
\urldef\tempurl%
\url{http://proceedings.mlr.press/v84/tomczak18a.html}
\showURL{%
\tempurl}


\bibitem[Vahdat and Kautz(2020)]%
        {ArashNVAE}
\bibfield{author}{\bibinfo{person}{Arash Vahdat} {and} \bibinfo{person}{Jan
  Kautz}.} \bibinfo{year}{2020}\natexlab{}.
\newblock \showarticletitle{NVAE: A Deep Hierarchical Variational Autoencoder}.
  In \bibinfo{booktitle}{\emph{Advances in Neural Information Processing
  Systems}}, \bibfield{editor}{\bibinfo{person}{H.~Larochelle},
  \bibinfo{person}{M.~Ranzato}, \bibinfo{person}{R.~Hadsell},
  \bibinfo{person}{M.~F. Balcan}, {and} \bibinfo{person}{H.~Lin}} (Eds.),
  Vol.~\bibinfo{volume}{33}. \bibinfo{publisher}{Curran Associates, Inc.},
  \bibinfo{pages}{19667--19679}.
\newblock
\urldef\tempurl%
\url{https://proceedings.neurips.cc/paper/2020/file/e3b21256183cf7c2c7a66be163579d37-Paper.pdf}
\showURL{%
\tempurl}


\bibitem[van~den Oord et~al\mbox{.}(2017)]%
        {OordVAE}
\bibfield{author}{\bibinfo{person}{Aaron van~den Oord}, \bibinfo{person}{Oriol
  Vinyals}, {and} \bibinfo{person}{koray kavukcuoglu}.}
  \bibinfo{year}{2017}\natexlab{}.
\newblock \showarticletitle{Neural Discrete Representation Learning}. In
  \bibinfo{booktitle}{\emph{Advances in Neural Information Processing
  Systems}}, \bibfield{editor}{\bibinfo{person}{I.~Guyon},
  \bibinfo{person}{U.~V. Luxburg}, \bibinfo{person}{S.~Bengio},
  \bibinfo{person}{H.~Wallach}, \bibinfo{person}{R.~Fergus},
  \bibinfo{person}{S.~Vishwanathan}, {and} \bibinfo{person}{R.~Garnett}}
  (Eds.), Vol.~\bibinfo{volume}{30}. \bibinfo{publisher}{Curran Associates,
  Inc.}
\newblock
\urldef\tempurl%
\url{https://proceedings.neurips.cc/paper/2017/file/7a98af17e63a0ac09ce2e96d03992fbc-Paper.pdf}
\showURL{%
\tempurl}


\bibitem[Vanegas et~al\mbox{.}(2009)]%
        {vanegas2009visualization}
\bibfield{author}{\bibinfo{person}{Carlos~A Vanegas}, \bibinfo{person}{Daniel~G
  Aliaga}, \bibinfo{person}{Bedrich Benes}, {and} \bibinfo{person}{Paul
  Waddell}.} \bibinfo{year}{2009}\natexlab{}.
\newblock \showarticletitle{Visualization of simulated urban spaces: Inferring
  parameterized generation of streets, parcels, and aerial imagery}.
\newblock \bibinfo{journal}{\emph{IEEE Transactions on Visualization and
  Computer Graphics}} \bibinfo{volume}{15}, \bibinfo{number}{3}
  (\bibinfo{year}{2009}), \bibinfo{pages}{424--435}.
\newblock


\bibitem[Vanegas et~al\mbox{.}(2010)]%
        {vanegas2010modelling}
\bibfield{author}{\bibinfo{person}{Carlos~A Vanegas}, \bibinfo{person}{Daniel~G
  Aliaga}, \bibinfo{person}{Peter Wonka}, \bibinfo{person}{Pascal M{\"u}ller},
  \bibinfo{person}{Paul Waddell}, {and} \bibinfo{person}{Benjamin Watson}.}
  \bibinfo{year}{2010}\natexlab{}.
\newblock \showarticletitle{Modelling the appearance and behaviour of urban
  spaces}. In \bibinfo{booktitle}{\emph{Computer Graphics Forum}},
  Vol.~\bibinfo{volume}{29}. Wiley Online Library, \bibinfo{pages}{25--42}.
\newblock


\bibitem[Vanegas et~al\mbox{.}(2012)]%
        {vanegas2012inverse}
\bibfield{author}{\bibinfo{person}{Carlos~A Vanegas}, \bibinfo{person}{Ignacio
  Garcia-Dorado}, \bibinfo{person}{Daniel~G Aliaga}, \bibinfo{person}{Bedrich
  Benes}, {and} \bibinfo{person}{Paul Waddell}.}
  \bibinfo{year}{2012}\natexlab{}.
\newblock \showarticletitle{Inverse design of urban procedural models}.
\newblock \bibinfo{journal}{\emph{ACM Transactions on Graphics (TOG)}}
  \bibinfo{volume}{31}, \bibinfo{number}{6} (\bibinfo{year}{2012}),
  \bibinfo{pages}{1--11}.
\newblock


\bibitem[Vaswani et~al\mbox{.}(2017)]%
        {VaswaniAttention}
\bibfield{author}{\bibinfo{person}{Ashish Vaswani}, \bibinfo{person}{Noam
  Shazeer}, \bibinfo{person}{Niki Parmar}, \bibinfo{person}{Jakob Uszkoreit},
  \bibinfo{person}{Llion Jones}, \bibinfo{person}{Aidan~N Gomez},
  \bibinfo{person}{\L~ukasz Kaiser}, {and} \bibinfo{person}{Illia Polosukhin}.}
  \bibinfo{year}{2017}\natexlab{}.
\newblock \showarticletitle{Attention is All you Need}. In
  \bibinfo{booktitle}{\emph{Advances in Neural Information Processing
  Systems}}, \bibfield{editor}{\bibinfo{person}{I.~Guyon},
  \bibinfo{person}{U.~V. Luxburg}, \bibinfo{person}{S.~Bengio},
  \bibinfo{person}{H.~Wallach}, \bibinfo{person}{R.~Fergus},
  \bibinfo{person}{S.~Vishwanathan}, {and} \bibinfo{person}{R.~Garnett}}
  (Eds.), Vol.~\bibinfo{volume}{30}. \bibinfo{publisher}{Curran Associates,
  Inc.}
\newblock
\urldef\tempurl%
\url{https://proceedings.neurips.cc/paper/2017/file/3f5ee243547dee91fbd053c1c4a845aa-Paper.pdf}
\showURL{%
\tempurl}


\bibitem[Vyas et~al\mbox{.}(2020)]%
        {vyas_et_al_2020}
\bibfield{author}{\bibinfo{person}{A. Vyas}, \bibinfo{person}{A.
  Katharopoulos}, {and} \bibinfo{person}{F. Fleuret}.}
  \bibinfo{year}{2020}\natexlab{}.
\newblock \showarticletitle{Fast Transformers with Clustered Attention}.
\newblock  (\bibinfo{year}{2020}).
\newblock


\bibitem[Wang et~al\mbox{.}(2020)]%
        {wang2020sceneformer}
\bibfield{author}{\bibinfo{person}{Xinpeng Wang}, \bibinfo{person}{Chandan
  Yeshwanth}, {and} \bibinfo{person}{Matthias Nie{\ss}ner}.}
  \bibinfo{year}{2020}\natexlab{}.
\newblock \showarticletitle{SceneFormer: Indoor Scene Generation with
  Transformers}.
\newblock \bibinfo{journal}{\emph{arXiv preprint arXiv:2012.09793}}
  (\bibinfo{year}{2020}).
\newblock


\bibitem[Weber et~al\mbox{.}(2009)]%
        {weber2009interactive}
\bibfield{author}{\bibinfo{person}{Basil Weber}, \bibinfo{person}{Pascal
  M{\"u}ller}, \bibinfo{person}{Peter Wonka}, {and} \bibinfo{person}{Markus
  Gross}.} \bibinfo{year}{2009}\natexlab{}.
\newblock \showarticletitle{Interactive geometric simulation of 4d cities}. In
  \bibinfo{booktitle}{\emph{Computer Graphics Forum}},
  Vol.~\bibinfo{volume}{28}. Wiley Online Library, \bibinfo{pages}{481--492}.
\newblock


\bibitem[Williams and Headleand(2017)]%
        {williams2017time}
\bibfield{author}{\bibinfo{person}{Benjamin Williams} {and}
  \bibinfo{person}{Christopher~J Headleand}.} \bibinfo{year}{2017}\natexlab{}.
\newblock \showarticletitle{A time-line approach for the generation of
  simulated settlements}. In \bibinfo{booktitle}{\emph{2017 International
  Conference on Cyberworlds (CW)}}. IEEE, \bibinfo{pages}{134--141}.
\newblock


\bibitem[Yang et~al\mbox{.}(2013)]%
        {yang2013urban}
\bibfield{author}{\bibinfo{person}{Yong-Liang Yang}, \bibinfo{person}{Jun
  Wang}, \bibinfo{person}{Etienne Vouga}, {and} \bibinfo{person}{Peter Wonka}.}
  \bibinfo{year}{2013}\natexlab{}.
\newblock \showarticletitle{Urban pattern: Layout design by hierarchical domain
  splitting}.
\newblock \bibinfo{journal}{\emph{ACM Transactions on Graphics (TOG)}}
  \bibinfo{volume}{32}, \bibinfo{number}{6} (\bibinfo{year}{2013}),
  \bibinfo{pages}{1--12}.
\newblock


\bibitem[You et~al\mbox{.}(2018)]%
        {you2018graphrnn}
\bibfield{author}{\bibinfo{person}{Jiaxuan You}, \bibinfo{person}{Rex Ying},
  \bibinfo{person}{Xiang Ren}, \bibinfo{person}{William~L. Hamilton}, {and}
  \bibinfo{person}{Jure Leskovec}.} \bibinfo{year}{2018}\natexlab{}.
\newblock \showarticletitle{GraphRNN: Generating Realistic Graphs with Deep
  Auto-regressive Models}. In \bibinfo{booktitle}{\emph{Proceedings of the 35th
  International Conference on Machine Learning, {ICML} 2018,
  Stockholmsm{\"{a}}ssan, Stockholm, Sweden, July 10-15, 2018}}
  \emph{(\bibinfo{series}{Proceedings of Machine Learning Research},
  Vol.~\bibinfo{volume}{80})}, \bibfield{editor}{\bibinfo{person}{Jennifer~G.
  Dy} {and} \bibinfo{person}{Andreas Krause}} (Eds.).
  \bibinfo{publisher}{{PMLR}}, \bibinfo{pages}{5694--5703}.
\newblock
\urldef\tempurl%
\url{http://proceedings.mlr.press/v80/you18a.html}
\showURL{%
\tempurl}


\bibitem[Zang and Wang(2020)]%
        {moflow}
\bibfield{author}{\bibinfo{person}{Chengxi Zang} {and} \bibinfo{person}{Fei
  Wang}.} \bibinfo{year}{2020}\natexlab{}.
\newblock \showarticletitle{MoFlow: An Invertible Flow Model for Generating
  Molecular Graphs}. In \bibinfo{booktitle}{\emph{Proceedings of the 26th ACM
  SIGKDD International Conference on Knowledge Discovery; Data Mining}}
  (Virtual Event, CA, USA) \emph{(\bibinfo{series}{KDD '20})}.
  \bibinfo{publisher}{Association for Computing Machinery},
  \bibinfo{address}{New York, NY, USA}, \bibinfo{pages}{617–626}.
\newblock
\showISBNx{9781450379984}
\urldef\tempurl%
\url{https://doi.org/10.1145/3394486.3403104}
\showDOI{\tempurl}


\end{thebibliography}


\appendix
\section{Background}
There exist many complementary frameworks for generative models -  Variational AutoEncoders (VAEs) \cite{KingmaWellingVAE}, Generative Adversarial Networks (GANs) \cite{GoodfellowGAN}, Energy Based Models (EBMs) \cite{AckleyBoltzmann}. All of these models can be built using modern building blocks of deep networks - Convolutional Neural Nets (CNNs) \cite{LecunnGradient}, transformers \cite{VaswaniAttention} and their numerous variants like UNET \cite{OlafUNET}, ResNet \cite{he2016deep} for CNNs and Axial Attention \cite{ho2020axial}, Longformer \cite{Beltagy2020LongformerTL} for Transformers. \\

\cpara{VAE} The standard VAE \cite{KingmaWellingVAE} is an encoder-decoder framework, where the data is encoded into latent variables, and decoded from those latents. By making sure that the latents come from a tractable distribution (usually a zero-mean, unit variance Gaussian), efficient sampling is achieved. Further refinements to this base include more expressive priors \cite{TomczakVampPrior, KingmaIAF, chen2016variational}, or better architectures \cite{ArashNVAE, Maale2019BIVAAV, Ranganath2016HierarchicalVM, Snderby2016LadderVA} which are often hierarchical. A complementary strategy to improve performance is to break down training into two steps: 1: autoencoding without any prior; 2: post-hoc learning of the latent. This is the approach adopted by \cite{Ghosh2020From, OordVAE, OordVAE2}.

Specifically, in \cite{OordVAE} the Vector Quantized VAE (VQVAE) is proposed. During training, the latents from the encoder are replaced by their closest element from a learned dictionary, a process known as Vector Quantization in signal processing literature \cite{GrayVQ}. As this dictionary has discrete indices, we are therefore able to learn a discrete representation of the input. A prior is then learned over this representation. \cite{OordVAE2} use a hierarchical version of the VQVAE to model high quality images of up to $1024 \times 1024$ pixels. \\

\cpara{Transformers} The transformer architecture \cite{VaswaniAttention} originally arose in the field of Natural Language Processing, but has since seen prolific use in fields as diverse as Object Detection \cite{CarionDETR, beal2020toward}, Image Classification \cite{dosovitskiy2021an}, Image Generation \cite{parmar2018image, esser2020taming} and Layout Generation \cite{para2020generative, wang2020sceneformer}. A transformer is composed of multiple layers of Attention Blocks \cite{parikh-etal-2016-decomposable, lin2017structured} which model dependencies of different positions in the sequence relative to each other. Transformers use sequences as their input. Concretely, if we have a sequence $\mathbf{x} = (x_1, x_2, \ldots, x_n)$ of length $n$, we embed these discrete tokens to get the features $X \in \mathbb{R}^{n \times d}$, where $d$ is the dimensionality of the embedding. From this, we compute representations via three independent MLPs and generate query $Q \in \mathbb{R}^{n \times d_q}$, key $K \in \mathbb{R}^{n \times d_q}$ and value, $V \in \mathbb{R}^{n \times d_v}$. The output of the block is then given by 
\[
    \text{Attention}(Q, K, V) = \text{softmax}\left(\frac{QK^T}{\sqrt{d_q}}\right) V
\]

The $n \times n$ matrix, $a_{ij} = \text{softmax}(QK^T)$ defines the interaction strength between each possible pairs of positions, and uses those as weights to combines the features from $V$. A transformer uses multiple layers of this operation. The matrix formed by the operation $QK^T$ has a size $n \times n$ which makes the attention/transformer architecture have an $n^2$ dependency on the length of the input sequence. A lot of research has therefore gone into reducing this dependency by using alternative attention schemes. \\

\cpara{Auto-Regressive Modelling} VAEs focus on decoding latents which are sampled from tractable distributions into entire data points, whether language, images or audio. An alternative to this is to use auto-regressive modelling where the likelihood of data point, say $\mathbf{x} = (x_1, x_2, \ldots, x_n)$ can be decomposed into the product of conditional likelihoods.
\[
p(\mathbf{x}) = \prod_{i=1}^{n} p(x_i | x_1, x_2, \ldots, x_{i-1})
\]

This is a \textit{causal} ordering meaning that the probability of the current position depends only on the previously generated positions. This is natural for modalities like language, where words inherently exist in a sequence, but somewhat non-intuitive for images. A simple way to treat the image as a sequence is to flatten the entire $n \times n$ image into a length $n^2$ sequence. This is one of the approaches used in \cite{OordRecurrent}. In images, where local context is generally more important, a simple modification is possible to save on computation. This modified factorization can be written as 

\[
p(\mathbf{x}) = \prod_{i=1}^{n} p(x_i | x_{<i})
\]
where $x_{<i}$ refers to any arbitrary subset of pixels \textit{before} the current pixel. This is usually chosen to be a $k \times k$ causal window around the current pixel which can easily be achieved by Convolutional Networks with proper masking \cite{Salimans2017PixeCNN, OordPixelCNN, jain2020lmconv}.

In the context of transformers, the so-called attention matrix, $a_{ij} = \text{softmax}(QK^T)$ breaks the causal-ordering by having any position depend on all positions in the sequence. This is fixed by setting $a_{ij} = 0$ for the elements where $ j \leq i$. With this, the transformer model can only model interactions between elements in the sequence before the current position.

We make heavy use of the VQVAE architecture and an Auto-Regressive transformer in our proposed method. 

\section{Priority Classes} \label{appendix:street_types}
We use data from the \osm to obtain information about the type of every street. In detail, we extract key-value pairs which are assigned to each \emph{way}, where a way is what the \osm denotes as a polyline, in our case representing a street section. The key highlighting a way to be a street is the term \emph{highway}, the value indicates the street type. Note that we do not consider street types like pedestrian ways or street types that do not really contribute to the actual street network of a city like service roads. We assign street sections to the defined priority classes $P_1$ and $P_2$ according to the values in Table \ref{tab:street_types}.

\begin{table}[h]
\centering
\begin{tabular}{l|lll}
$P_1$                   & $P_2$                  \\ \cline{1-2}
\emph{motorway}         & \emph{tertiary}        \\
\emph{trunk}            & \emph{tertiary\_link}  \\
\emph{primary}          & \emph{residential}  &  \\
\emph{secondary}        & \emph{living\_street}  \\
\emph{motorway\_link}   & \emph{road}            \\
\emph{trunk\_link}      & \emph{unclassified}    \\
\emph{primary\_link}    &                        \\
\emph{secondary\_link}  &

\end{tabular}
\caption{Assignment of \osm street types to our three priority classes.}
\label{tab:street_types}
\end{table}


\section{L-System streets}

Table ~\ref{tab:ce_params} contains the short and long street length parameters used to generate the street networks. These values were sampled directly from the area in the ground-truth, by computing each block boundary, finding the minimum bounding box, and reporting the short or long length of the box. We took range within one standard deviation of the mean. Following~\cite{Parish:2001:PMC} we used the implementation of CityEngine. An obstacle map was used to mark the water. Following typical street design in North America, major streets followed an \emph{organic} pattern, and minor streets follow a \emph{raster} pattern. 50,000 streets were generated and cropped to the region extent. The implementation allowed no control over raster orientation or whether a region bounded by major streets would be filled with minor streets.

\begin{table}
\centering
\begin{tabular}{ c | c | c | c | c }
& width (m) & deviation & length (m) & deviation \\
\hline
Washington & 93.63  & 93.75 & 185.48 & 186.87\\
Chicago & 123.91 & 96.35 & 255.29 & 182.69 \\
Los-Angeles & 104.26 & 86.28 & 213.58 & 178.88 \\
San Francisco & 96.81 & 75.98 & 196.63 & 159.82 \\
New York & 92.97 & 78.47 & 182.51 & 156.60  \\
\end{tabular}
\caption{Parameters sampled from the ground-truth used on the L-System streets.}
\label{tab:ce_params}
\end{table}

\section{Statistical measures}

Here we describe the statistical measures used in the evaluation of our networks, but refer the interested reader to ~\cite{boeing2020multi} as an example of the wide area graph statistics in use by the urban planning community. 

A \emph{segment} is a sequence of edges in the graph without branches, connected by vertices with valencies only of 2. \emph{circuity} measures the ratio between the length of the segment (the sum of the length of the edges in the segment) against the euclidean distance between the first and last vertex of the segment. It describes the local curvature of the streets.

The \emph{transport ratio}~\cite{boeing2019morphology} measures the ratio between the walking distance (along the graph) and the euclidean between two vertices. We sample its value over $n=500$ randomly chosen vertex pairs in the graph. The transport distance describes the connectivity of the city - how easy it is to get between nearby points.

\emph{Pagerank} is an algorithm originally intended to measure the connectivity of webpages~\cite{page1999pagerank}. It has since been applied to the study of street networks~\cite{jiang2009ranking}, providing a measure of topological connectivity. It models the movements of a traveler through the graph starting from a random vertex, stepping between adjacent vertices randomly, continuing with a probability $k=0.85$ (and halting with a probability $1-k$). As is only a measure of topological connectivity, ignoring vertex position and edge length. To address this, we introduce \emph{Pagerank-by-edge}. In this model, the traveler travels over an edge graph (EG). An EG has a vertex for every edge in the input graph, and an edge to all adjacent edges in the input graph. We use a $k=0.95$ to simulate longer walks on large graphs. It highlights areas with long edges and high connectivity, at the cost of over-penalizing short edges.

\begin{figure*}[t]
    \centering
    \includegraphics[width=0.95\textwidth]{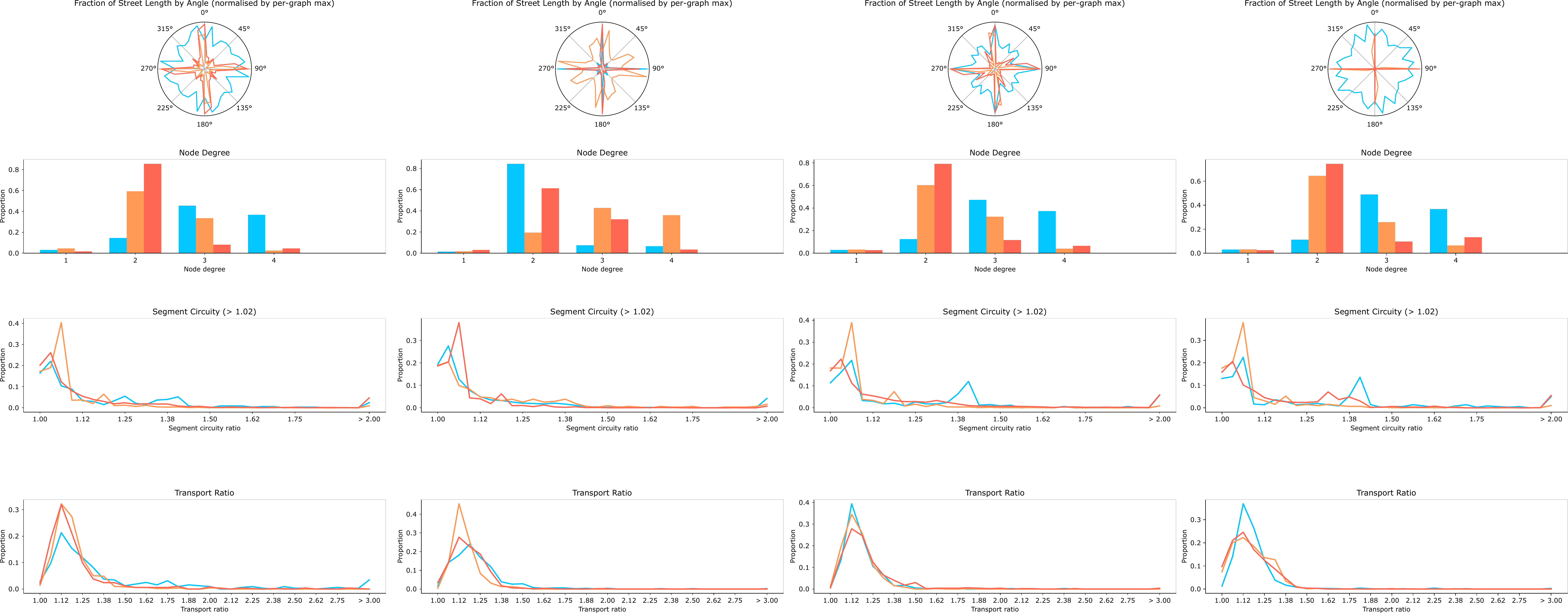}
    \caption{We show statistics from the ground-truth in Red, our model in Orange, and CityEngine in Blue. The cities from left-to-right are Chicago, San Francisco, Washington DC and Los Angeles.}
    \label{fig:all_stats}
\end{figure*}

The reported density values normalize the values over the area of land in the street graph. We show all of these statistics in Fig. \ref{fig:all_stats}.


\begin{figure*}[b]
    \centering
    \includegraphics[width=0.95\textwidth]{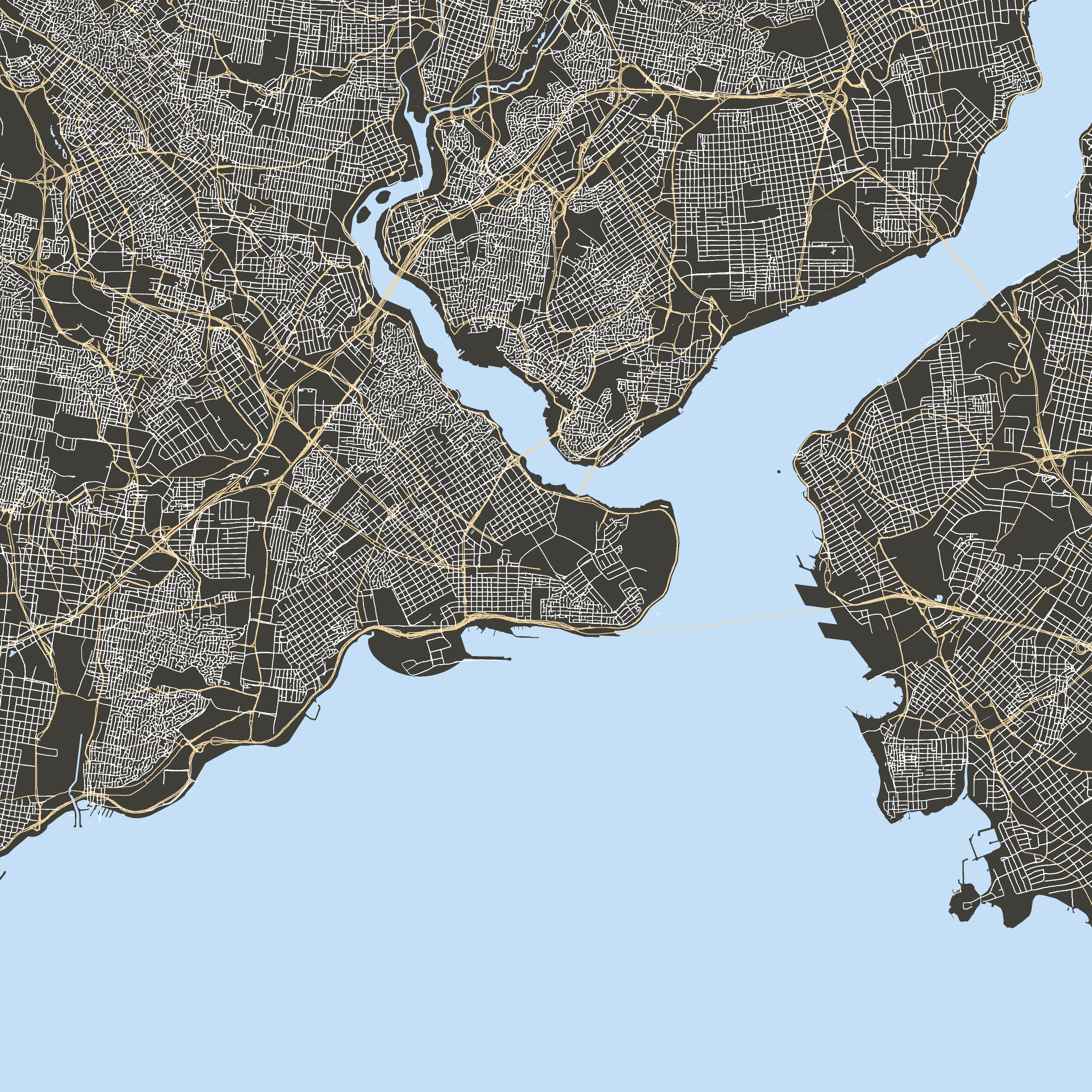}
    \caption{Alternative street layout for the city of Istanbul.}
    \label{fig:istanbul-large}
\end{figure*}


\end{document}